%% file: main.tex
\documentclass[10pt,twocolumn,letterpaper]{article}

\usepackage[pagenumbers]{iccv}
\usepackage{nopageno}

\input{preamble}

\definecolor{iccvblue}{rgb}{0.21,0.49,0.74}
\usepackage[pagebackref,breaklinks,colorlinks,allcolors=iccvblue]{hyperref}
\usepackage{booktabs}
\usepackage{multirow}
\usepackage{multicol}
\usepackage[mathscr]{euscript}
\usepackage[dvipsnames]{xcolor}
\usepackage{colortbl}
\usepackage{makecell}
\usepackage{subcaption}
\usepackage[symbol]{footmisc}
\usepackage{kotex}
\usepackage{colortbl}

\newcommand{\et}[2]{${#1}^{\pm{#2}}$}
\newcommand{\etb}[2]{$\mathbf{{#1}}^{\pm{#2}}$}
\newcommand{\ets}[2]{$\underline{{#1}}^{\pm{#2}}$}
\def\paperID{11129} % *** Enter the Paper ID here
\def\confName{ICCV}
\def\confYear{2025}

\title{DisCoRD: \underline{Dis}crete Tokens to \underline{Co}ntinuous Motion via \underline{R}ectified Flow \underline{D}ecoding}

\def\spaces{~~~~~~}

\author{Jungbin Cho\textsuperscript{1}\thanks{Equal contribution. 
}
\spaces{}Junwan Kim\textsuperscript{1}\footnotemark[1]\spaces{}Jisoo Kim\textsuperscript{1}\spaces{}Minseo Kim\textsuperscript{1}\spaces{}Mingu Kang\textsuperscript{2}\\\spaces{}Sungeun Hong\textsuperscript{2}\spaces{}Tae-Hyun Oh\textsuperscript{1,3}\spaces{}Youngjae Yu$^{1,3}$\\\\
\textsuperscript{1}Yonsei University\\
\textsuperscript{2}Sungkyunkwan University\\
\textsuperscript{3}POSTECH
} 

\let\oldtwocolumn\twocolumn
\renewcommand\twocolumn[1][]{%
    \oldtwocolumn[{#1}{
    \begin{center}
            \includegraphics[width=0.98\textwidth]{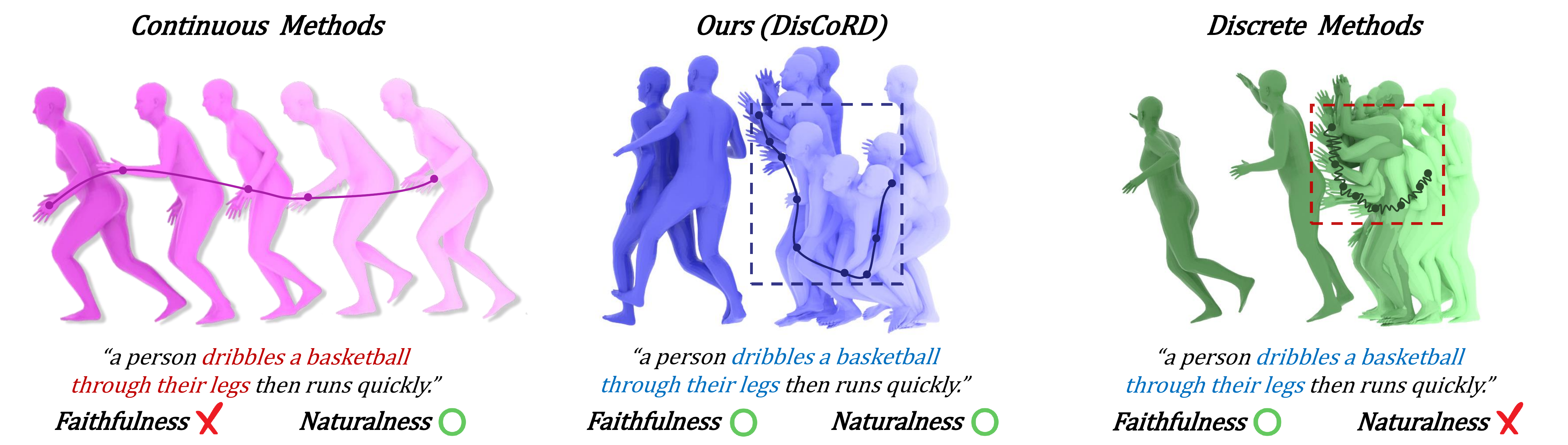}
            
            \vspace{-6pt}
            \captionof{figure}{{\color{Magenta}Continuous methods} generate smooth motions, but lack faithfulness (red text) to conditioning signals. In contrast, {\color{ForestGreen}discrete methods} demonstrate high faithfulness (blue words) but often produce less natural results such as unexpressive motion and frame-wise noise (red box). We present a novel discrete token decoding method, {\color{Violet}\textbf{DisCoRD}}, that generates smooth, dynamic motion (blue box) while faithfully adhering to the conditioning signal. The plotted lines represent left-hand trajectories of generated motions for visual comparison.}

            \label{fig:teaser}
        \end{center}
    }]
}

\begin{document}
\maketitle
\input{sec/0_abstract}    
\input{sec/1_introduction}

\input{sec/2_RelatedWorks}

\begin{figure*}[t] \centering
    \includegraphics[width=\textwidth]{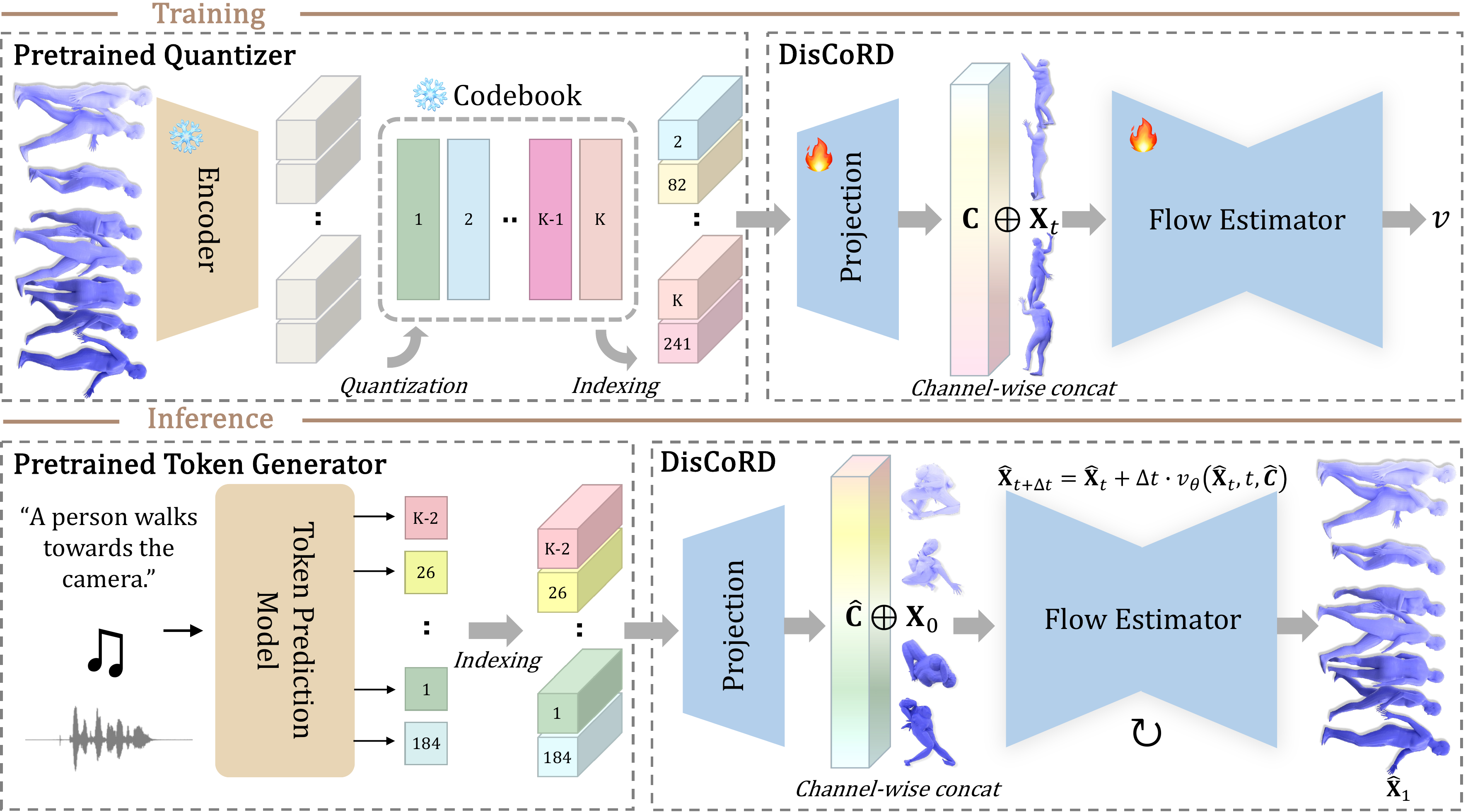}
    \caption{\textbf{An overview of DisCoRD.} During the \textbf{Training} stage, we leverage a pretrained quantizer to first obtain discrete representations (tokens) of motion. These tokens are then projected into continuous features $\mathbf{C}$, which are concatenated with noisy motion $\mathbf{X}_t$. This concatenated feature is used to train a vector field $v$. During the \textbf{Inference} stage, we use a pretrained token prediction model based on the pretrained quantizer to first generate tokens from the given control signal. These generated tokens are then projected into continuous features $\mathbf{\hat C}$, concatenated with Gaussian noise $\mathbf{X}_0\sim \mathcal{N}(0,I)$, and iteratively decoded through the learned vector field $v_\theta$ into motion $\mathbf{\hat X}_1$.}
    % \vspace{-1.3em}
\label{fig:overall_pipeline}
\end{figure*}

\input{sec/3_Method}

\input{sec/4_Experiments}
\input{sec/5_Conclusion}
{
    \small
    \bibliographystyle{ieeenat_fullname}
    \bibliography{main}
}
% {
%     \small
%     \input{main.bbl}
% }

\let\twocolumn\oldtwocolumn
\input{sec/X_suppl} 

\end{document}

%% file: preamble.tex
\usepackage{microtype}
\renewcommand{\paragraph}[1]{\vspace{1mm}\noindent\textbf{#1}}

%% file: sec/0_abstract.tex
\begin{abstract}
Human motion is inherently continuous and dynamic, posing significant challenges for generative models. While discrete generation methods are widely used, they suffer from limited expressiveness and frame-wise noise artifacts. In contrast, continuous approaches produce smoother, more natural motion but often struggle to adhere to conditioning signals due to high-dimensional complexity and limited training data.
To resolve this ``discord'' between discrete and continuous representations we introduce \textbf{DisCoRD}: Discrete Tokens to Continuous Motion via Rectified Flow Decoding, a novel method that leverages rectified flow to decode discrete motion tokens in the continuous, raw motion space. Our core idea is to frame token decoding as a conditional generation task, ensuring that DisCoRD captures fine-grained dynamics and achieves  smoother, more natural motions. Compatible with any discrete-based framework, our method enhances naturalness without compromising faithfulness to the conditioning signals on diverse settings. Extensive evaluations demonstrate that DisCoRD achieves state-of-the-art performance, with FID of 0.032 on HumanML3D and 0.169 on KIT-ML. These results establish DisCoRD as a robust solution for bridging the divide between discrete efficiency and continuous realism. Project website: \href{https://whwjdqls.github.io/discord-motion/}{https://whwjdqls.github.io/discord-motion/}
\end{abstract}

%% file: sec/1_introduction.tex
\vspace{-1.3em}

\section{Introduction}
\label{sec:Introduction}

Human motion generation controlled by diverse signals has become an emerging area in computer vision, driven by its vast applications in virtual reality to animation, gaming, and human-computer interaction. The ability to generate realistic human motions that are precisely aligned with input conditions—such as textual descriptions \cite{humanmotion, guo2024momask, chen2023mld, guo_generating_3dhuman}, human speech \cite{Talkshow, emage, probtalk}, or even music \cite{aist++, siyao2022bailando, gong2023tm2d}—is essential for creating immersive and interactive experiences. Two critical qualities define the success of such systems \cite{motionBERT_fidcannotcatch}: \textit{faithfulness}, ensuring that the generated motion accurately reflects the conditioning signal, and \textit{naturalness}, producing smooth and lifelike motions that are comfortable and convincing to human observers. 
Deficiencies in faithfulness cause misaligned motions, while slight unnaturalness disrupts immersion and triggers the uncanny valley \cite{uncanyvalley} effect.
% A deficit in faithfulness can produce motions that are misaligned or irrelevant to the intended context, while even slight unnaturalness can disrupt user immersion, creating an uncanny valley effect \cite{uncanyvalley}.

%     이런 차이가 FID랑 MDS로 measure된다까지 넣으면 너무 길어지는듯..?} \label{fig:intro}
% \end{figure}

% To achieve both naturalness and faithfulness in human motion generation, recent methods employ a two-stage approach. A primary strategy involves continuous latent space techniques, where a motion Variational Autoencoder (VAE) is trained to establish a smooth, continuous representation \cite{chen2023mld}. This continuous latent space aligns well with the smooth dynamics of human motion, offering a basis for generating natural movements. However, while these methods are effective at generating visually coherent motions, they often encounter challenges with cross-modal mapping ambiguity, where a single input signal can correspond to multiple plausible motions. This issue becomes especially pronounced in data-constrained settings, such as motion capture datasets, where limited examples exacerbate the difficulty of learning consistent mappings between signals and motions.
Since human motion is inherently continuous, generation based on continuous representations is naturally well-suited for producing smooth and realistic motion \cite{zhang2022motiondiffuse, humanmotion, chen2023mld, Actor}. However, due to the high dimensionality of continuous representation, they often encounter challenges with cross-modal mapping ambiguity \cite{codetalker, discreteactionsarebetter} which can result in low faithfulness. This issue becomes especially pronounced in data-constrained settings, such as motion capture datasets \cite{amass}, where limited examples lead to difficulty of learning consistent mappings between signals and motions. On the other hand, discrete quantization methods \cite{guo2024momask, pinyoanuntapong2024mmm, zhang2023t2m} utilize motion VQ-VAEs \cite{vqvae} to discretize motion representation, simplifying the learning of high-dimensional data mappings by reformulating it as a classification task. This discretization enables more efficient learning and can be particularly beneficial when dealing with limited data, improving faithfulness \cite{guo2024momask, pinyoanuntapong2024bamm, pinyoanuntapong2024mmm}. However, motion VQ-VAEs face two main challenges. 
% First, under-reconstruction occurs when token discretization removes fine-grained motion details, which are crucial for generating dynamic movements. 
First, under-reconstruction occurs when fine-grained motion details, which are essential for generating dynamic movements, are lost during token discretization.
% Second, frame-wise noise arises when discrete tokens are directly decoded, introducing unnatural artifacts that disrupt motion smoothness and reduce user immersion. 
Second, frame-wise noise arises from directly decoding discrete tokens, introducing unnatural artifacts that disrupt motion smoothness and diminish user immersion.
These challenges make it difficult to generate motion that is both smooth and natural while maintaining high faithfulness.

% both high fidelity and natural motion generation.

% under-reconstruction, caused by representing continuous motion with a finite set of tokens, and frame-wise noise, caused when decoding these discrete representations. Under-reconstruction can lead to loss of fine-grained motion details, essential for dynamic motion, and frame-wise noise can introduce unnatural artifacts, degrading the naturalness of generated motions and diminishing user immersion.

% Due to the two  issues in reconstruction of motions in stage one issues not only affect the reconstruction of motions in stage one, but also propagate into the second stage of training, degrading the naturalness of the generated motions.

% While ensuring naturalness is essential, traditional metrics like MPJPE do not align well with human perception \cite{review3dhumanpose_mpjpe_notgood}, and FID falls short in capturing fine-grained, frame-wise noise \cite{EDGE_fidcannotcatch, motionBERT_fidcannotcatch, motionpercept2024_fidcannotcatch}. This limitation reduces the effectiveness of these metrics in assessing naturalness, particularly in cases where under-reconstruction and frame-wise noise co-exists. To address this, we introduce a novel evaluation scheme, the Motion Deviation Score (MDS), designed to evaluate reconstructed motion by ensuring it is free from under-reconstructed areas while also minimizing fine-grained noise. 

% Our experiments demonstrate the effectiveness of DisCoRD in enhancing sample-wise naturalness without encountering issues of under-reconstruction.

\begin{figure}[t] \centering
    \includegraphics[width=0.45\textwidth]{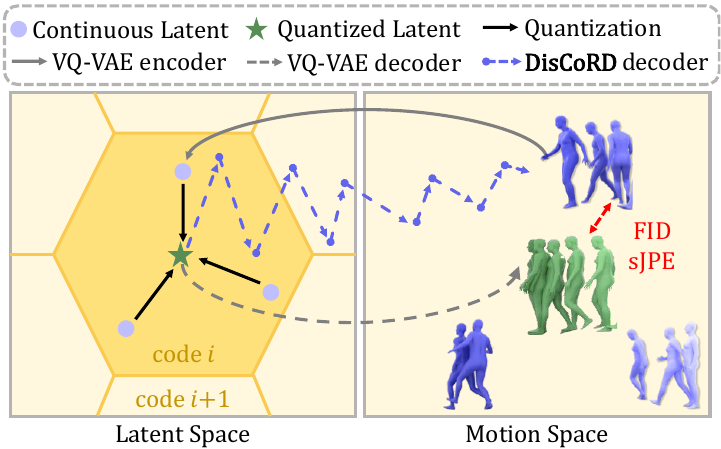}
    \caption{\textbf{Concept of DisCoRD.} Discrete quantization methods encode multiple motions into a single quantized representation. While existing methods directly decode from this quantized representation, DisCoRD iteratively decodes the discrete latent in a continuous space to recover the inherent continuity and dynamism of motion. To assess the gap between reconstructed and real motion, prior work primarily used FID as the metric. Here, we additionally propose symmetric Jerk Percentage Error (sJPE) to evaluate the differences in naturalness between reconstructed and real motion.}
    \vspace{-1.3em}
\label{fig:intro}
\end{figure}

In this paper, we propose \textbf{DisCoRD}, a novel approach that bridges discrete and continuous motion generation to achieve both faithfulness and naturalness. 
% Our key insight is to enhance the strong faithfulness of discrete methods by leveraging rectified flow models to decode pretrained discrete tokens into raw motion space. 
% This process improves motion smoothness while maintaining precise alignment with conditioning signals (Figure~\ref{fig:teaser}).
Our key insight is to leverage the strong faithfulness of discrete generation methods \cite{zhang2023t2m, guo2024momask, pinyoanuntapong2024bamm} by utilizing rectified flow models \cite{liu2022flow, lipman2022flow} to translate pretrained discrete tokens back into raw motion space. This enhances naturalness while preserving alignment with the conditioning signals.

Our method offers two primary advantages over traditional discrete decoding methods, as shown in Figure~\ref{fig:intro}. First, instead of directly decoding discrete tokens into motion space, we use them as conditioning signals to guide motion generation within the continuous motion space. This reduces fine-grained noise and results in smoother, more natural motion. Second, rather than relying on a one-step decoding process, we employ an iterative refinement approach using a rectified flow model~\cite{birodkar2024sample, zhao2024epsilon}, which progressively improves reconstruction quality. This enables the generation of dynamic and complex movements that conventional methods struggle to capture. Moreover, \textbf{DisCoRD} is framework-agnostic, making it adaptable to any discrete generation method (e.g., autoregressive~\cite{zhang2023t2m} or bidirectional~\cite{guo2024momask}), regardless of the conditioning signal type (e.g., text, music, or speech), thereby improving performance.
% Moreover, \textbf{DisCoRD} is framework-agnostic, making it adaptable to any discrete-based motion generation methods (e.g autoregressive or bidirectional), regardless of its conditioning signal (e.g text, music, or speech), to improve performance.

% Its flexibility enables seamless integration into existing methods, various tasks.

% While our method improves motion naturalness, evaluating the naturalness of reconstructed motions remains challenging as traditional metrics like MPJPE do not align well with human perception \cite{review3dhumanpose_mpjpe_notgood, human36m}, and FID fails to capture subtle, frame-wise noise, shown in Figure~\ref{fig:fid_verses_mds}. This limitation reduces the effectiveness of these metrics in assessing the naturalness of reconstructed motions. 
Although our method improves motion naturalness, evaluating this quality remains challenging. Traditional metrics such as MPJPE do not correlate well with human perception \cite{review3dhumanpose_mpjpe_notgood, human36m}, and FID fails to capture subtle, frame-wise noise, as illustrated in Figure~\ref{fig:fid_verses_mds}. These limitations reduce the reliability of existing metrics in accurately assessing reconstructed motion naturalness.
% particularly in cases of under-reconstruction and frame-wise noise. 
% To address this, we introduce a novel metric: the symmetric Jerk Percentage Error (\textbf{sJPE}), designed to evaluate reconstructed motion by simultaneously detecting under-reconstruction and fine-grained artifacts, per sample.
To address this, we introduce a novel sample-wise metric, the symmetric Jerk Percentage Error (\textbf{sJPE}), which evaluates reconstructed motion by simultaneously detecting under-reconstruction and fine-grained artifacts.
% To address this, we introduce a novel metric: the symmetric Jerk Percentage Error (\textbf{sJPE}), designed to evaluate reconstructed motion by ensuring it is free from under-reconstruction while also minimizing fine-grained noise. 
Our experiments demonstrate the effectiveness of DisCoRD in enhancing sample-wise naturalness without suffering from under-reconstruction.
Extensive evaluations across text-to-motion, co-speech gesture, and music-to-dance generation demonstrate that our method achieves state-of-the-art naturalness, consistently outperforming existing approaches. Our contributions are as follows:
% Additionally, our method is adaptable to any discrete-based motion generation framework, independent of conditioning signals, to improve performance. Extensive experiments across diverse tasks—text-to-motion generation, co-speech gesture generation, and music-to-dance generation—demonstrate that our approach integrates seamlessly with various existing methods, achieving state-of-the-art naturalness in most cases according to both standard evaluation metrics and our proposed metric, sJPE. 
\begin{enumerate}
    \item We introduce \textbf{DisCoRD}, a novel method for decoding discrete tokens in continuous motion space, improving the naturalness of generated motions while preserving faithfulness across various models and tasks.
    \item We propose a novel evaluation scheme, the symmetric Jerk Percentage Error (\textbf{sJPE}), designed to evaluate both under-reconstruction and frame-wise noise, which are often overlooked but critical for motion generation.
    \item Our extensive experiments demonstrate that our methods achieve state-of-the-art performance on existing human motion generation scenarios.
\end{enumerate}
\vspace{-0.5em}
% , we found that traditional ground truth-based metrics are insufficient to assess physical naturalness in motion. To address this, we introduce an additional metric, Jitter \cite{yi2021transpose}, as an effective measure of physical naturalness. Our analysis on jitter reveals that while conventional methods may perform well on standard metrics, they often lack physical coherence. In contrast, our method not only excels in conventional metrics but also generates more natural and physically consistent motion.

% Overall, our contributions are as follows: 1) We achieve state-of-the-art performance across most existing human motion generation scenarios. 2) We address the critical, often-overlooked problem of generating natural human motion, which discrete quantization methods frequently fail to capture. 3) We propose a novel framework to enhance the naturalness of human motion in continuous space, compatible with any discrete-based method. We conduct extensive experiments across various motion generation tasks to verify that our method is a general solution for improving the naturalness of discrete quantization based motion generation frameworks.

%% file: sec/2_RelatedWorks.tex
\section{Related Work}
\label{sec:Relatedwork}
\vspace{-0.5em}
\paragraph{Human Motion Generation from Diverse Signals.} Generating natural and controllable 3D human motion remains a long-standing task. Early approaches prioritized motion naturalness~\cite{motiongraph, motionfromexample, parametricmotiongraphs}, while recent advances in deep learning expand capabilities to generate motion conditioned on diverse signals. Recent advances of text-to-motion datasets \cite{guo_generating_3dhuman, kit} have driven progress in text-conditioned motion synthesis, while music-motion datasets \cite{aist++, deepdance} have enabled music-driven dance generation. Speech-motion datasets \cite{Talkshow, beat, emage} extends motion control by synthesizing gestures from speech. As most recent methods rely on discrete representations~\cite{guo2024momask, gong2023tm2d, probtalk}, our work enhances motion naturalness for all discrete methods, regardless of the task, by addressing the discrete token decoding problem.

% The goal of human motion generation is to generate natural motions that adhere to control signals. Recent advancements in this field can be broadly categorized into two main approaches: motion generation from continuous representations, where models directly regress on continuous values, and motion generation from discrete representations, where motion is quantized into discrete tokens, transforming the generation task into a classification problem in the discrete domain.

\paragraph{Continuous Human Motion Generation.}  Early approaches to signal-to-motion generation  employ regression-based mapping of control signals to motion within a continuous representation space. These works leverage Variational Autoencoders (VAE) \cite{kingma2013auto, guo_generating_3dhuman, petrovich2022temos}, GANs \cite{gan_1}, or CLIP features \cite{radford2021learning, tevet2022motionclip} to generate natural motion. More recently, with the success of diffusion models \cite{ho2020denoising, song2020score}, motion generation models have achieved unprecedented generation quality \cite{tevet2023human, zhang2022motiondiffuse, EDGE_fidcannotcatch}. 
Follow-up works explore continuous latent spaces for efficient motion generation \cite{chen2023mld, motionLCM}, incorporate physical constraints to improve realism \cite{yuan2023physdiff}, and integrate retrieval mechanisms to enhance generalization \cite{zhang2023remodiffuse}. 
% These approaches leverage continuous representations to better capture the inherent properties of motion, 
Their ability to generate smooth, natural motion 
makes them well-suited for not only motion synthesis but also for a generative prior \cite{priormdm, buddi}. However, due to the lack of scalability in current motion datasets, the high complexity of continuous representations often makes it difficult to establish reliable cross-modal mappings between control signals and generated motions, leading to suboptimal performance.

\paragraph{Discrete Human Motion Generation.}
Recently, to simplify the complex mapping between control signals and motions, some methods have reformulated the generation task as a discrete token classification problem, achieving notable performance in motion generation \cite{zhang2023t2m, Talkshow, siyao2022bailando}.
% \cite{pinyoanuntapong2024mmm, zhang2023t2m, guo2022tm2t, guo2024momask, pinyoanuntapong2024bamm, hosseyni2024bad, , yang2024unimumo, liu2024towards, Talkshow, emage}. 
These approaches often employ VQ-VAEs \cite{van2017neural} and its variants \cite{RQVAE} to create motion tokens, which are then used to generate motion sequences via autoregressive \cite{zhang2023t2m, pinyoanuntapong2024bamm, hosseyni2024bad,Talkshow, synergy, motionllm}, or masked \cite{guo2024momask, intermask, emage, probtalk} token prediction. 
% token prediction models, such as autoregressive transformers. 
More recently, discrete diffusion models have been introduced to directly denoise these discrete tokens \cite{m2d2m, lou2023diversemotion}. While these methods effectively bypass complex signal-to-motion mapping challenges, their inherent characteristics—such as quantization errors and discreteness result in unnatural artifacts, including under-reconstruction and frame-wise noise.

%% file: sec/3_Method.tex
\section{Method}

In this section, we introduce DisCoRD, a novel method for decoding pretrained discrete representations in the raw motion domain using rectified flow. This approach enables motion generation that is both smooth and dynamic: (1) decoding in the raw motion domain preserves natural motion smoothness, and (2) utilizing rectified flow enhances expressiveness, capturing fast-paced movements. We begin by introducing rectified flow models and motion tokenization, followed by an explanation of our condition projection, conditional rectified flow decoder, and training details.

\subsection{Preliminaries}

\paragraph{Rectified Flow.} Diffusion models~\cite{ho2020denoising,song2020score} have demonstrated remarkable performance due to their iterative denoising formulation, which enhances their ability to capture complex data variations and generate high-dimensional samples. However, they typically require a large number of denoising steps to produce high-quality outputs. In contrast, rectified flow~\cite{liu2022flow} provides a more direct approach by framing sample generation as a transport problem, addressed through a flow matching algorithm~\cite{liu2022flow, albergo2022building, lipman2022flow}.  Flow matching algorithm aims to construct a transport map denoted as $T: \mathbb{R}^d\rightarrow\mathbb{R}^d$, that effectively transfers observations from the source distribution $x_0 \sim \pi_0 $ on $\mathbb{R}^d$ to the target distribution $x_1 \sim \pi_1$ on $\mathbb{R}^d$. This transport process is formalized as the following ordinary differential equation (ODE):
\begin{equation}
    dx_t = v(x_t, t) \, dt.
\label{eq:fm_ode}
\end{equation}
Here, $v$ represents the vector field, and $x_t$ denotes the trajectory parameterized over $t \in [0,1]$. Rectified flow follows the formulation of the forward process in diffusion models \cite{kingma2024understanding}, but its specific parameterization enables a more direct mapping between distributions and improves efficiency. Specifically, its forward process can be expressed as:
\begin{equation}
    x_t = tx_1 + (1 - t)x_0,
\label{eq:rf}
\end{equation}
where $v$ is defined as $x_1 - x_0$. Then, the model is trained to learn a causal approximation of $v$, denoted as $v_\theta$, by solving the following least squares regression problem:
\begin{equation}
    \begin{gathered}
        \min_v \int_0^1 \mathbb{E} \left[ \left\| (x_1 - x_0) - v(x_t, t) \right\|^2 \right] \, dt.
    \end{gathered}
    % \vspace{-0.5em}
\label{eq:flow_loss}
\end{equation}
Once trained, samples from the target distribution $\pi_1$ can be generated by solving Equation~\eqref{eq:fm_ode} using an ODE solver, where the initial conditions are drawn from the source distribution $\pi_0$. Unlike conventional diffusion models, which require many denoising steps, rectified flow follows a nearly straight trajectory, enabling more efficient transport with much fewer denoising steps.

\paragraph{Motion Tokenization.} The objective of generative models based on discrete quantization methods is to reformulate the regression problem into a classification problem. These models typically undergo a two-stage training process. In stage 1, a VQ-VAE is trained to encode a motion sequence \( \mathbf{X} = [\mathbf{x}_1, \mathbf{x}_2, \dots, \mathbf{x}_T] \) where \( \mathbf{x}_t \in \mathbb{R}^{d_{motion}} \), using an encoder \( \mathcal{E} \), into a sequence of discrete tokens \( \mathbf{Z} = [\mathbf{z}_1, \mathbf{z}_2, \dots, \mathbf{z}_{T/q}] \). Each token \( \mathbf{z}_t \) is retrieved from the codebook \( \mathcal{Z}=\left\{\mathbf{z}_k \in \mathbb{R}^{d_{code}} \right\}_{k=1}^{N} \), and \( T \) represents the length of the original motion sequence while \( q \) is the downsample factor. Then a decoder \( \mathcal{D} \) reconstructs the motions \( \mathbf{X}_{recon} \) from \( \mathbf{Z} \), with the network trained using a reconstruction loss and a commitment loss. In stage 2, the index sequence \( \mathbf{S} = [\mathbf{s}_1, \mathbf{s}_2, \dots, \mathbf{s}_{T/q}] \), representing the one-hot encoded codebook indices of the discrete token sequence \( \mathbf{Z} \), is used to train a next-index prediction model conditioned on various signals.

After training both stages, generating a new motion sequence \( \hat{\mathbf{X}} \) from a given condition \( \mathbf{C} \) involves two steps: first, the stage 2 model produces a sequence of predicted indices \( \hat{\mathbf{S}} \), which is then converted into discrete tokens \( \hat{\mathbf{Z}} \) using the learned codebook. Finally, \( \mathcal{D} \) reconstructs the motion sequence \( \hat{\mathbf{X}} \) from \( \hat{\mathbf{Z}} \), yielding the desired motion output.

\subsection{DisCoRD}
% Traditional deterministic feed-forward decoders $\mathcal{D}$ 
Directly decoding discrete tokens using traditional feed-forward decoders $\mathcal{D}$ suffer from limited expressiveness and propagate token discreteness into decoded motions, resulting in under-reconstructed and noisy outputs (Figure~\ref{fig:sjpe_comparison}). To address these issues, we propose decoding pretrained tokens in continuous space by replacing $\mathcal{D}$ with an expressive rectified flow model. Specifically, we first extract frame-wise conditioning features from discrete tokens through a Condition Projection module, and then use these features as frame-wise conditions for a Rectified Flow Decoder that synthesizes human motion from Gaussian noise. The overall pipeline of DisCoRD is depicted in Figure~\ref{fig:overall_pipeline}.
\paragraph{Condition Projection.} To enable our decoder to generate expressive motion, we first extract frame-wise conditioning features from discrete tokens \( \mathbf{Z} = [\mathbf{z}_1, \dots, \mathbf{z}_{T/q}] \). 
Since each token $\mathbf{z}_t$ encodes information spanning $q$ consecutive motion frames, we must extract $q$ distinct, frame-specific features from each token.
% A na\"ive upsampling and linear projection would and its associated motion frames and upconvolution layers would disregard the temporal correspondence between each token .
A na\"ive upsampling and linear projection would result in same $q$ features from each token, and upconvolution layers would disregard the temporal correspondence between each tokens and frames.
To mitigate these issues, we first repeat each token \( \bold{z}_t \in \mathbb{R}^{1 \times d_{code}}\) \( q \) times to restore the original temporal resolution, resulting in \(\bold{z}^{repeat}_t \in\mathbb{R}^{q \times d_{code}}\). Then we stack into a tensor \( \mathbf{z}_{t}^{stacked} \in \mathbb{R}^{1 \times (q \times d_{code})} \) and project to \(\bold{z}^{project}_t \in \mathbb{R}^{1 \times (q \times d_{feat})}\). Finally, we unstack the projected tensor to $\bold{z}^{final}_t \in \mathbb{R}^{q \times d_{feat}}$ where each vector $\mathbf{c}_{i} \in \mathbb{R}^{d_{feat}} (\text{for}\ i\in[1,...,q])$ are frame-wise conditioning features. This process is applied to every token in \( \mathbf{Z} \), resulting in \(\mathbf{C} = [\mathbf{c}_1, \dots, \mathbf{c}_{T}] \). This approach maintains the correspondence between tokens and motion frames by explicitly extracting $q$ features from each token, ensuring that the resulting frame-wise conditioning features are well-suited for the motion decoding. Moreover, we found that our projection method enhances motion generation on unseen token sequences on stage 2.

% To mitigate these issues, each token \( \bold{z}_t \in \mathbb{R}^{1 \times d_{code}}\) is first repeated \( q \) times to restore the original temporal resolution, resulting in \(\bold{z}^{repeat}_t \in\mathbb{R}^{q \times d_{code}}\). Then we stack into a tensor \( \mathbf{z}_{t}^{stacked} \in \mathbb{R}^{1 \times (q \times d_{code})} \) and project to \(\bold{z}^{project}_t \in \mathbb{R}^{1 \times (q \times d_{feat})}\). 

% 이방법론은 correspondence를 preserve하기 때문에 feature들이 frame-wise conditioning을 하기 적합핟. 

% effectively guiding the iterative decoding process to yield high quality motion.

% 우린 continuous space에서 디코딩 할꺼다
% 그래서 token을 가이드로만 써서 raw motion space에서 decoding을했다. 
\paragraph{Rectified Flow Decoder.} 
Our goal is to decode discrete tokens into natural motion while operating within a continuous space. Therefore, we do not directly map discrete tokens back into motion, but treat discrete tokens as a signal to guide motion decoding in the raw motion space. 
Given the frame-wise conditioning features $\mathbf{C}$ extracted from discrete tokens by the condition projection module, we train a conditional rectified flow model to reconstruct the original motion. 
% Unlike conventional methods that directly map discrete tokens back into motion, our approach treats discrete tokens as guiding signals rather than explicit motion representations. 
% By leveraging frame-wise conditioning features $\mathbf{C}$ extracted from discrete tokens, we enable the decoding process to occur in a smooth and expressive continuous space. 
Specifically, given a motion data distribution $P_\mathbf{X}$, we define the transport process from Gaussian noise $\mathbf{X_0}\sim\mathcal{N}(0,I)$ to motion $\mathbf{X_1}\sim P_{\mathbf{X}}$, guided by frame-wise conditioning features $\mathbf{C}$, formulated as: 
% which are features extracted from discrete tokens, formulated as:
\begin{equation}
    \begin{gathered}
        \min_v \int_0^1 \mathbb{E} \left[ \left\| (\mathbf{X}_1 - \mathbf{X}_0) - v(\mathbf{X}_t, t, \mathbf{C}) \right\|^2 \right] \, dt, \\
        \text{with} \quad \mathbf{X}_t = t \mathbf{X}_1 + (1 - t) \mathbf{X}_0.
    \end{gathered}
    % \vspace{-0.5em}
\label{eq:flow_loss_cond}
\end{equation}
% our goal is to decode discrete codebooks into raw motion using dense, frame-wise conditions. 
% To implement this, 
We concatenate frame-wise features $\mathbf{C}$ along the channel dimension, similar to image generation methods \cite{zhao2024epsilon, ho2021cascadeddiffusionmodelshigh}, allowing each motion frame to be conditioned independently.
This formulation ensures that decoding remains in the continuous space, enabling more expressive decoding.
% This setup allows each motion frames to be conditioned on per-frame codebook features, enabling more expressive decoding. conditional ODE using the Euler method. 
% In contrast to traditional diffusion models that generate motion from sparse conditions like text or action labels \cite{humanmotion}, our goal is to decode discrete codebooks into raw motion using dense, frame-wise conditions. We achieve this by concatenating frame-wise features $\mathbf{C}$ extracted from discrete tokens along the channel dimension, similar to image generation methods \cite{zhao2024epsilon, ho2021cascadeddiffusionmodelshigh}. This setup allows each motion frames to be conditioned on per-frame codebook features, enabling more expressive decoding.
During inference, features $\mathbf{\hat{C}}$ are first extracted from tokens generated by a pretrained token generation model. This extracted features are then iteratively decoded into $\mathbf{\hat{X}_1}$ by solving a conditional ODE using the Euler method, progressively improving generation quality.

\paragraph{Training. } In variants of diffusion models, training is performed over the entire data instance with fixed sequence lengths, such as 196 frames in HumanML3D \cite{guo_generating_3dhuman}, a standard in motion generation \cite{humanmotion,chen2023mld}. While this approach improves reconstruction quality in stage 1, our results indicate that these improvements do not translate effectively to generation quality in stage 2. To address this limitation and improve generalization to unseen motion sequences during inference, we train our rectified flow model on sliding windows of motion frames rather than max length spans. Additionally, although conventional U-Net diffusion models often incorporate attention mechanisms to enhance performance \cite{latentdiffusionmodel}, we found this strategy to be suboptimal in our context, resulting in a performance degradation.

%% file: sec/4_Experiments.tex
\section{Experiments}
\label{sec:Experiments}
In this section, we evaluate the effectiveness of \textbf{DisCoRD} in achieving motion naturalness compared to other discrete methods. We begin by assessing the naturalness of reconstructed motions to highlight the expressive capabilities of our rectified flow decoder. Then, we examine how this naturalness carries over to stage 2, generating natural motions while preserving faithfulness. We focus on text-to-motion generation due to its complex motions and diversity, but also evaluate our approach on other motion generation tasks, including co-speech gesture generation and music-to-dance generation, demonstrating the flexibility of our method.

\subsection{Dataset and Evaluation}
\textbf{Datasets.} For text-to-motion, we use HumanML3D \cite{guo_generating_3dhuman} and KIT-ML \cite{kit}. \textbf{HumanML3D} is a 3D motion dataset with language annotations, including 14,616 motion sequences paired with 44,970 text descriptions. Sourced from motion capture data, motions are standardized to a template, scaled to 20 FPS, and cropped to 10 seconds if longer. 
% Each sequence has at least three descriptive texts covering diverse actions. 
\textbf{KIT-ML} is a smaller dataset with 3,911 motion sequences paired with 6,278 text descriptions. Motion capture data are downsampled to 12.5 FPS, with 1–4 descriptions per sequence. For co-speech gesture generation, we utilize the \textbf{SHOW} \cite{Talkshow} dataset, while a mixed version of \textbf{AIST++} \cite{aist++} and HumanML3D is used for music-to-dance generation. Further dataset details are provided in the Supplementary Section B.
% Section~\ref{sec:datasets_and_evaluations}.

% \begin{figure}[thb] \centering
%     \includegraphics[width=0.48\textwidth]{example-image}
%     \caption{what is under-recon, and noise? graph and qualitative} \label{fig:under_recon_over_recon}
% \end{figure}
 % given its complexity and the extensive scale of available text-to-motion datasets. 
\begin{figure}[t] \centering
    \includegraphics[width=0.5\textwidth]{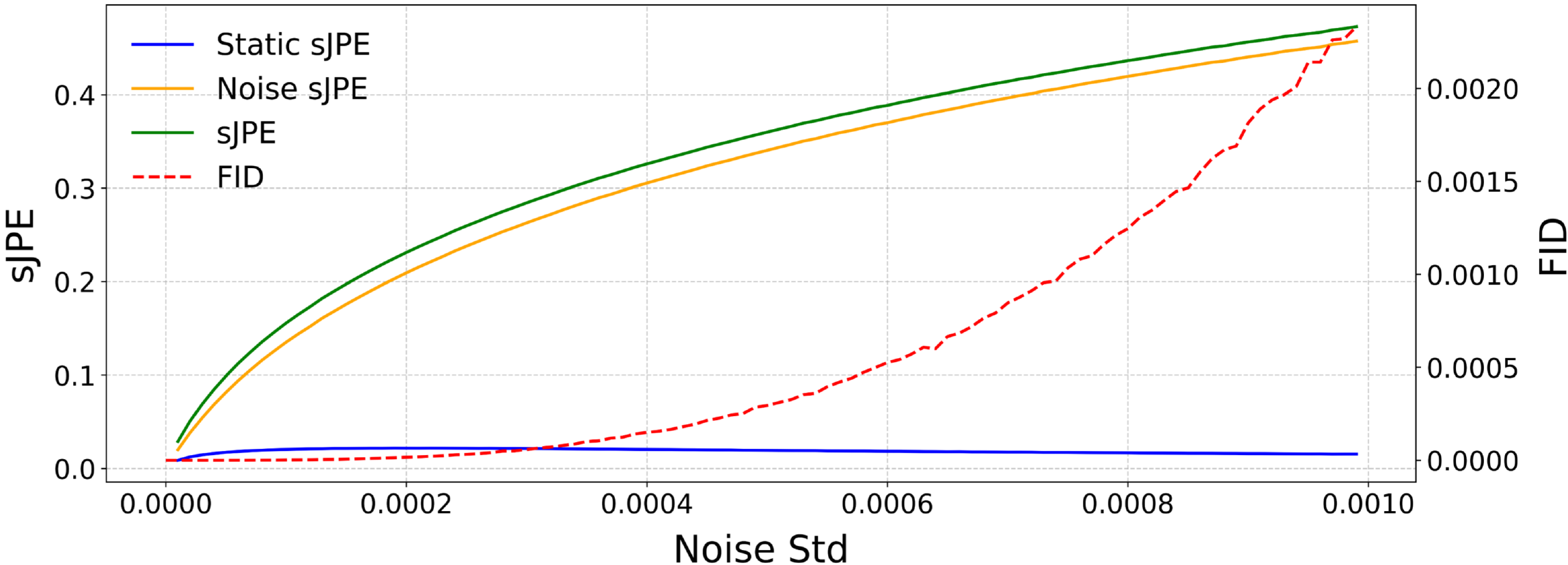}
\caption{\textbf{sJPE and FID response to frame-wise gaussian noise.} We introduce Gaussian noise with varying standard deviations (x-axis) to ground-truth motion data and evaluate its effect on sJPE and FID. \textit{Noise sJPE} is highly sensitive to subtle frame-wise perturbations, whereas \textit{Static sJPE} remains low. FID is highly insensitive to frame-wise noise. Note that FID scale (y-axis, right) is very small compared to sJPE scale (y-axis, left).}
\label{fig:fid_verses_mds}
    \vspace{-0.5em}

\end{figure}

\begin{figure*}[t] \centering
    \includegraphics[width=0.99\textwidth]{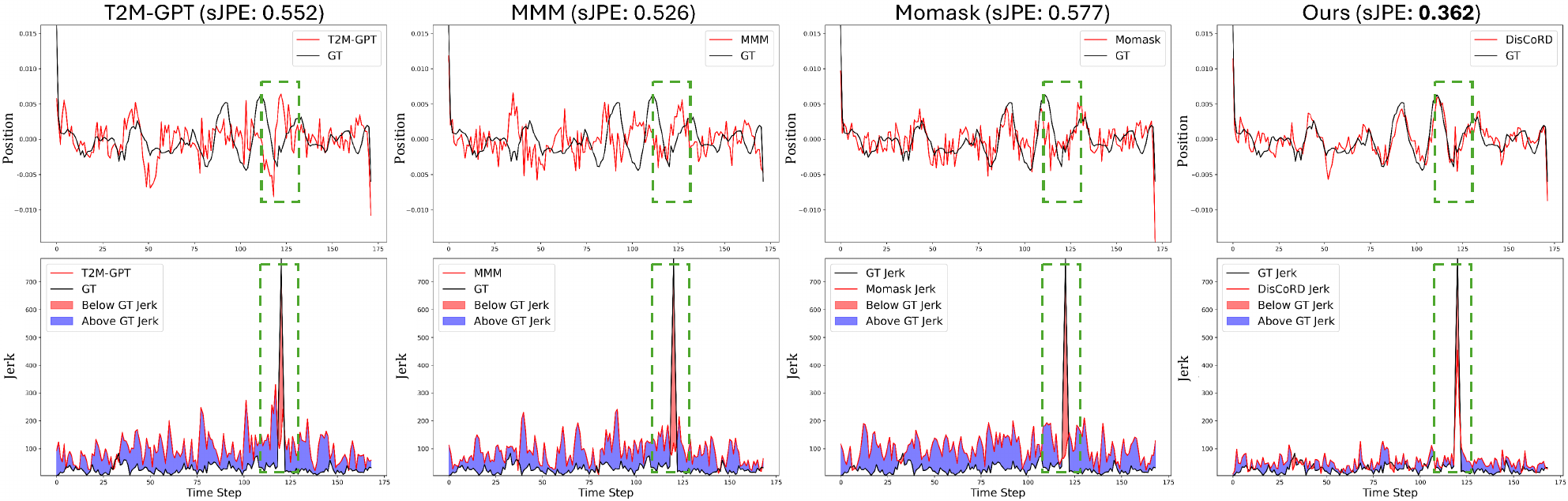}
    \caption{\textbf{Under-reconstruction and frame-wise noise.} We visualize fine-grained motion trajectories (top), and corresponding jerk graphs (bottom), where blue and red regions indicate noise and static sJPE, respectively. Compared to other methods, DisCoRD significantly reduces sJPE, resulting in smoother motion (fewer blue regions) and greater dynamism (fewer red regions), as highlighted in green boxes.}
    % . Compared to different methods, DisCoRD demonstrates lower noise levels (Noise sJPE, blue area) and reduced under-reconstruction (Static sJPE, red area), with under-reconstruction regions in Momask highlighted in green boxes. This results in a lower sJPE for our method, indicating improved naturalness in the reconstructed motion.} 
    \label{fig:sjpe_comparison}
    % \vspace{-1.3em}
\end{figure*}

\paragraph{Evaluation.} We evaluate DisCoRD on both motion reconstruction and motion generation separately. For motion reconstruction, the primary objective is to assess how effectively the decoder reconstructs motion from tokens. This is measured by Fréchet Inception Distance (FID), which assesses motion realism by comparing the feature distributions of generated and ground truth motions, and Mean Per Joint Position Error (MPJPE), which quantifies positional accuracy. 
For text-to-motion generation, we follow \cite{humanmotion} and employ several established metrics: FID, R-Precision, Multimodal Distance (MM-Dist), and Multimodality (MModality).
% For text-to-motion generation, we follow prior works \cite{humanmotion} and we employ several established metrics: FID; R-Precision, which measures retrieval accuracy by matching generated motions to corresponding text descriptions and reports Top-1, Top-2, and Top-3 accuracies; Multimodal Distance (MM-Dist), which calculates the average feature distance between generated motions and associated text descriptions, capturing cross-modal alignment; and Multimodality (MModality), which assesses the model's capacity to produce diverse motions from a single text prompt by averaging distances between multiple generations for the same prompt. 
For co-speech gesture generation, we employ Fréchet Gesture Distance (FGD) \cite{FGD}, and for music-to-dance generation, following \cite{EDGE_fidcannotcatch}, we utilize Dist$_k$ and Dist$_g$ to assess the quality of generated motions by comparing the distributional spread of generated and real motions. Additional specifics on these metrics are provided in the Supplementary Section B.

\paragraph{Symmetric Jerk Percentage Error.} Prior works \cite{chen2023mld, guo2024momask} rely on MPJPE and FID at stage 1. However, MPJPE has limited correlation with human perceptual preferences \cite{review3dhumanpose_mpjpe_notgood}, while FID, being a model-level metric extracted from pretrained network features, fails to capture per-sample naturalness \cite{motionpercept2024_fidcannotcatch}. Our experiments further indicate that FID is particularly insensitive to subtle, fine-grained noise (see Figure \ref{fig:fid_verses_mds}), which critically affects immersive motion quality \cite{noiseavatar}. To overcome these limitations, we introduce the Symmetric Jerk Percentage Error (SJPE), a metric explicitly designed to assess both under-reconstructed motions and frame-level noise through jerk. Jerk, defined as the third derivative of position with respect to time, has proven to effectively quantify subtle deviations \cite{FlowMDM, m2d2m} and kinetic inconsistencies in motion \cite{ jerk_1, jerk_2}, being a critical measure for detecting unnatural artifacts in generated motion.

Let \( J_{\text{pred},t} \) and \( J_{\text{true},t} \) denote the predicted and ground truth jerk, respectively, at time \( t \) over \( n \) time points. Then sJPE, capturing the symmetric mean absolute percentage error \cite{MAKRIDAKIS1993527} between predicted and ground truth jerk values, is defined as
\begin{equation}
    \text{sJPE} = \tfrac{1}{n} \sum\nolimits_{t=1}^{n} \tfrac{|J_{\text{pred},t} - J_{\text{true},t}|}{|J_{\text{true},t}| + |J_{\text{pred},t}|}.
\label{eq:mds}
\end{equation}
Within sJPE, we identify two components: \textit{Noise sJPE} and \textit{Static sJPE}. Noise sJPE corresponds to instances where \( J_{\text{pred},t} > J_{\text{true},t} \), indicating an overestimation of jerk in the predicted motion, which reflects the presence of fine-grained noise. This effect is evident in discrete-based methods, where discrete tokens introduce frame-wise noise, as shown in Figure~\ref{fig:sjpe_comparison}. Static sJPE captures instances where \( J_{\text{pred},t} \leq J_{\text{true},t} \), indicating underestimation of jerk, or insufficiently dynamic predicted motion, highlighted by green boxes in Figure~\ref{fig:sjpe_comparison}. Together, these components provide a comprehensive measure of prediction accuracy, capturing both over- and underestimations of jerk within a unified score. Additional details and results are in Supplementary Section C.

\begin{table}[t]
    \centering
    \scalebox{0.73}{
    \begin{tabular}{l l c c c}
    \toprule
    Dataset & Methods & \hphantom{000} FID $\downarrow$ \hphantom{000} & MPJPE $\downarrow$ & \hphantom{000}sJPE $\downarrow$\hphantom{000} \\
    \hline
    \addlinespace[0.3em] 
\multicolumn{1}{>{\columncolor{white}\arraybackslash}l}{\multirow{7}{*}
{\makecell[c]{\textcolor{black}{}\\ %
\textcolor{black}{H-ML3D}}}
}

        & MLD~\cite{chen2023mld} (cont.)  & 0.017 & 14.7 & 0.404 \\
          \cline{2-5}
          \addlinespace[0.3em] 
        & T2M-GPT~\cite{zhang2023t2m} & 0.089 & \textbf{60.0} & 0.564\\
        \rowcolor{blue!13}\cellcolor{white}
        & \textbf{+DisCoRD(Ours)} & \textbf{0.031}\makebox[0pt][l]{\color{blue}\scriptsize{(+65\%)}} & 71.5 & \textbf{0.488}\makebox[0pt][l]{\color{blue}\scriptsize{(+13\%)}} \\
           \cline{2-5}
           \addlinespace[0.3em] 
        & MMM~\cite{pinyoanuntapong2024mmm} & 0.097 & \textbf{46.9} & 0.517 \\
        \rowcolor{blue!13}\cellcolor{white}
        & \textbf{+DisCoRD(Ours)} & \textbf{0.020}\makebox[0pt][l]{\color{blue}\scriptsize{(+79\%)}} & 56.8 & \textbf{0.429}\makebox[0pt][l]{\color{blue}\scriptsize{(+17\%)}}\\
                \cline{2-5}
        \addlinespace[0.3em] 
        & MoMask~\cite{guo2024momask} & 0.019 & \textbf{29.5} & 0.512 \\
        \rowcolor{blue!13}\cellcolor{white}
        & \textbf{+DisCoRD(Ours)} & \textbf{0.011}\makebox[0pt][l]{\color{blue}\scriptsize{(+42\%)}} & 33.3 & \textbf{0.385}\makebox[0pt][l]{\color{blue}\scriptsize{(+25\%)}}\\
    \hline
    \midrule
    \multirow{4}{*}{\cellcolor{white}\makecell[c]{\\  KIT-ML}}
        & T2M-GPT~\cite{zhang2023t2m} & 0.470 & \textbf{46.4} & 0.526 \\
        \rowcolor{blue!13}\cellcolor{white}
        & \textbf{+DisCoRD(Ours)}  & \textbf{0.284}\makebox[0pt][l]{\color{blue}\scriptsize{(+40\%)}} & 58.7 & \textbf{0.395}\makebox[0pt][l]{\color{blue}\scriptsize{(+25\%)}} \\
        \cline{2-5}
         \addlinespace[0.3em] 
        & MoMask~\cite{guo2024momask} & 0.113 & 37.5 & 0.384 \\
        \rowcolor{blue!13}\cellcolor{white}
        & \textbf{+DisCoRD(Ours)} & \textbf{0.103}\makebox[0pt][l]{\color{blue}\scriptsize{(+9\%)}} & \textbf{33.0} & \textbf{0.359}\makebox[0pt][l]{\color{blue}\scriptsize{(+7\%)}}\\
    \bottomrule
    \end{tabular}
    }
\caption{\textbf{Quantitative results on motion reconstruction.} DisCoRD enhances naturalness as a decoder for discrete models, shown by improvements over base models on FID and sJPE (blue). H-ML3D stands for HumanML3D and cont. for continuous.}
\label{tab:stage1}
\vspace{-1em}
\end{table}

% Within sJPE, we identify two components: \textit{Noise sJPE} and \textit{Static sJPE}. Noise sJPE corresponds to instances where \( J_{\text{pred},t} > J_{\text{true},t} \), indicating an overestimation of jerk in the predicted motion, which reflects the presence of fine-grained noise. This effect is evident in discrete-based methods, where discrete tokens introduce frame-wise noise. Static sJPE captures instances where \( J_{\text{pred},t} \leq J_{\text{true},t} \), indicating underestimation of jerk, or insufficiently dynamic predicted motion. Together, these components provide a comprehensive measure of prediction accuracy, capturing both over- and underestimations of jerk within a unified score. In Figure~\ref{fig:sjpe_comparison}, we visualize the comparison of sJPE alongside joint position. The blue regions represent Noise sJPE, while the red regions correspond to Static sJPE. DisCoRD effectively mitigates both types of errors, significantly reducing both blue and red areas, which indicates smoother and more dynamic motion generation. This is further supported by the lower overall sJPE values, demonstrating DisCoRD’s ability to produce motion that closely aligns with the ground truth while minimizing frame-wise noise and under-reconstruction. More details and results are in Supplementary Section C.

% Section~\ref{sec:additional_analysis_on_sJPE} 
% for details on sJPE and qaulitatve results.

\vspace{0.1em}
\subsection{Quantitative results}
% Section~\ref{sec:datasets_and_evaluations}.
\vspace{0.1em}

\begin{table*}[t]
    \centering
    \scalebox{0.82}{
    \begin{tabular}{l l c c c c c c}
    \toprule
    \multirow{2}{*}{Datasets} & \multirow{2}{*}{Methods}  & \multicolumn{3}{c}{R Precision $\uparrow$} & \hphantom{0000} \multirow{2}{*}{FID $\downarrow$}\hphantom{000000} & \multirow{2}{*}{MM-Dist$\downarrow$} & \multirow{2}{*}{MultiModality $\uparrow$} \\
    \cline{3-5}
    \addlinespace[0.2em]  
    & & Top 1 & Top 2 & Top 3 \\
    \midrule
    \multirow{10}{*}{\makecell[c]{Human\\ML3D}} 
            % & Ground Truth &\et{0.511}{.003} &  \et{0.703}{.003} & \et{0.797}{.002} & \et{0.002}{.000} &\et{2.974}{.008} & - \\
            %    \cline{2-8}
            %     \addlinespace[0.3em]  
        & MDM~\cite{tevet2023human} & - & - & \et{0.611}{.007} & \et{0.544}{.044} & \et{5.566}{.027} & \etb{2.799}{.072} \\
        & MLD~\cite{chen2023executing} & \et{0.481}{.003} & \et{0.673}{.003} & \et{0.772}{.002} & \et{0.473}{.013} & \et{3.196}{.010} & \ets{2.413}{.079} \\
        & MotionDiffuse~\cite{zhang2022motiondiffuse} & \et{0.491}{.001} & \et{0.681}{.001} & \et{0.782}{.001} & \et{0.630}{.001} & \et{3.113}{.001} & \et{1.553}{.042} \\
        & ReMoDiffuse~\cite{zhang2023remodiffuse} & \et{0.510}{.005} & \et{0.698}{.006} & \et{0.795}{.004} & \et{0.103}{.004} & \et{2.974}{.016} & \et{1.795}{.043} \\
        & MMM~\cite{pinyoanuntapong2024mmm} & \et{0.504}{.003} & \et{0.696}{.003} & \et{0.794}{.002} & \et{0.080}{.003} & \et{2.998}{.007} & \et{1.164}{.041} \\
    \cline{2-8}
        \addlinespace[0.3em]  
        & T2M-GPT~\cite{zhang2023t2m} & \et{0.491}{.003} & \et{0.680}{.003} & \et{0.775}{.002} & \et{0.116}{.004} & \et{3.118}{.011} & \et{1.856}{.011} \\
        \rowcolor{blue!13}\cellcolor{white}
        & \textbf{+ DisCoRD (Ours)} & \et{0.476}{.008} & \et{0.663}{.006} & \et{0.760}{.007} & \et{0.095}{.011}\makebox[0pt][l]{\color{blue}\scriptsize{(+18\%)}} & \et{3.121}{.009} & \et{1.831}{.048} \\
    % \cline{2-8}
    %     \addlinespace[0.3em]  
    %     & MMM~\cite{pinyoanuntapong2024mmm} & \et{0.504}{.003} & \et{0.696}{.003} & \et{0.794}{.002} & \et{0.080}{.003} & \et{2.998}{.007} & \et{1.164}{.041} \\
    %     & \textbf{+ DisCoRD (Ours)} & \et{0.504}{.003} & \et{0.696}{.003} & \et{0.794}{.002} & \et{0.080}{.003} & \et{2.998}{.007} & \et{1.164}{.041} \\
    \cline{2-8}
        \addlinespace[0.3em]  
        & BAMM~\cite{pinyoanuntapong2024bamm} & \etb{0.525}{.002} & \etb{0.720}{.003} & \etb{0.814}{.003} & \et{0.055}{.002} & \etb{2.919}{.008} & \et{1.687}{.051} \\
        \rowcolor{blue!13}\cellcolor{white}
        & \textbf{+ DisCoRD (Ours)} & \et{0.522}{.003} & \ets{0.715}{.005} & \ets{0.811}{.004} & \ets{0.041}{.002}\makebox[0pt][l]{\color{blue} \scriptsize{(+25\%)}} & \ets{2.921}{.015} & \et{1.772}{.067} \\
    \cline{2-8}
        \addlinespace[0.3em]  
        & MoMask~\cite{guo2024momask} & \et{0.521}{.002} & \et{0.713}{.002} & \et{0.807}{.002} & \et{0.045}{.002} & \et{2.958}{.008} & \et{1.241}{.040} \\
        \rowcolor{blue!13}\cellcolor{white}
        & \textbf{+ DisCoRD (Ours)} & \ets{0.524}{.003} & \ets{0.715}{.003} & \et{0.809}{.002} & \etb{0.032}{.002}\makebox[0pt][l]{\color{blue} \scriptsize{(+29\%)}} & \et{2.938}{.010} & \et{1.288}{.043} \\
    \hline
    \midrule
    \multirow{10}{*}{\makecell[c]{KIT-\\ML}} 
            % & Ground Truth &\et{0.424}{.005} &  \et{0.649}{.006} & \et{0.779}{.006} & \et{0.031}{.004} &\et{2.788}{.012} & - \\
            %    \cline{2-8}
            %     \addlinespace[0.3em]  
        & MDM~\cite{tevet2023human} & - & - & \et{0.396}{.004} & \et{0.497}{.021} & \et{9.191}{.022} & \et{1.907}{.214} \\
        & MLD~\cite{chen2023executing} & \et{0.390}{.008} & \et{0.609}{.008} & \et{0.734}{.007} & \et{0.404}{.027} & \et{3.204}{.027} & \etb{2.192}{.071} \\
        & MotionDiffuse~\cite{zhang2022motiondiffuse} & \et{0.417}{.004} & \et{0.621}{.004} & \et{0.739}{.004} & \et{1.954}{.062} & \et{2.958}{.005} & \et{0.730}{.013} \\
        & ReMoDiffuse~\cite{zhang2023remodiffuse} & \et{0.427}{.014} & \et{0.641}{.004} & \et{0.765}{.055} & \etb{0.155}{.006} & \et{2.814}{.012} & \et{1.239}{.028} \\
        & MMM~\cite{pinyoanuntapong2024mmm} & \et{0.404}{.005} & \et{0.621}{.006} & \et{0.744}{.005} & \et{0.316}{.019} & \et{2.977}{.019} & \et{1.232}{.026} \\
    \cline{2-8}
        \addlinespace[0.3em]  
        & T2M-GPT~\cite{zhang2023t2m} & \et{0.398}{.007} & \et{0.606}{.006} & \et{0.729}{.005} & \et{0.718}{.038} & \et{3.076}{.028} & \et{1.887}{.050} \\
        \rowcolor{blue!13}\cellcolor{white}
        & \textbf{+ DisCoRD (Ours)} & \et{0.382}{.007} & \et{0.590}{.007} & \et{0.715}{.004} & \et{0.541}{.038}\makebox[0pt][l]{\color{blue} \scriptsize{(+25\%)}} & \et{3.260}{.028} & \ets{1.928}{.059} \\
    \cline{2-8}
        \addlinespace[0.3em]  
        & MoMask~\cite{guo2024momask} & \ets{0.433}{.007} & \ets{0.656}{.005} & \etb{0.781}{.005} & \et{0.204}{.011} & \etb{2.779}{.022} & \et{1.131}{.043} \\
        \rowcolor{blue!13}\cellcolor{white}
        & \textbf{+ DisCoRD (Ours)} & \etb{0.434}{.007} & \etb{0.657}{.005} & \ets{0.775}{.004} & \ets{0.169}{.010}\makebox[0pt][l]{\color{blue} \scriptsize{(+17\%)}} & \ets{2.792}{.015} & \et{1.266}{.046} \\
    \bottomrule
    \end{tabular}
    }
    \caption{\textbf{Quantitative results on motion generation.} $\pm$ indicates a 95\% confidence interval. +DisCoRD indicates that the baseline model’s decoder is replaced with DisCoRD. \textbf{Bold} indicates the best result, while \underline{underscore} refers the second best. DisCoRD improves naturalness, as evidenced by FID improvements shown in blue, while preserving faithfulness, demonstrated by R-Precision.} 
    % \vspace{-1.3em}
    \label{tab:quantitative_eval}
\end{table*}

\paragraph{Natural Motion Reconstruction.} We evaluate DisCoRD's effectiveness in reconstructing natural motions from discrete models. Existing discrete methods often struggle to generate natural motions, as indicated by higher sJPE and FID values. While MoMask achieves competitive FID, its high sJPE suggests significant frame-wise noise and poor reconstruction quality compared to continuous models like MLD. Our proposed DisCoRD decoder substantially improves motion quality, as shown by reduced FID and sJPE metrics, overcoming the typical limitations of discrete models and producing smoother, more natural motions.
Note that MPJPE measures only positional accuracy, and does not reflect motion naturalness or align with human perception \cite{human36m}.

\begin{table}[t]
    \centering
    \scalebox{0.85}{
    \begin{tabular}{l c c}
    \toprule
    Methods & sJPE$\downarrow$ & FGD$\downarrow$ \\
    \hline    
        \addlinespace[0.2em] 
        TalkSHOW~\cite{Talkshow} & 0.284 & 74.88 \\
        \rowcolor{blue!13}
        \textbf{+ DisCoRD(Ours)} & \textbf{0.077} & \textbf{43.58} \\
    \hline
        \addlinespace[0.2em] 
        ProbTalk~\cite{probtalk} & 0.406 & 5.21 \\
        \rowcolor{blue!13}
        \textbf{+ DisCoRD(Ours)} & \textbf{0.349} & \textbf{4.83} \\
    \bottomrule
    \end{tabular}
    }
    \caption{\textbf{Quantitative results on each method's SHOW test set.} DisCoRD outperforms baseline models on sJPE and FGD.}
    \label{tab:cospeech}
\end{table}

\begin{table}[t]
    \centering
    \scalebox{0.85}{
    \begin{tabular}{l c c c}
    \toprule
    Methods & sJPE$\downarrow$ & $\mathrm{Dist_k}\rightarrow$ (9.780) & $\mathrm{Dist_g}\rightarrow$ (7.662) \\
    \hline    
        \addlinespace[0.2em] 
        TM2D~\cite{gong2023tm2d} & 0.275 & 8.851 & 4.225 \\
        \rowcolor{blue!13}
        \textbf{+DisCoRD(Ours)} & \textbf{0.261} & \textbf{9.830} & \textbf{8.519} \\
    \bottomrule
    \end{tabular}
    }
    \caption{\textbf{Quantitative results on the AIST++ test set.} DisCoRD outperforms baseline model on sJPE, $\mathrm{Dist_k}$ and $\mathrm{Dist_g}$.}
    \vspace{-0.5em}
    \label{tab:dance}
\end{table}
% T2M-GPT uses vanilla VQ-VAE, which has limited representation power. In DisCoRD, a single token can map to multiple motions, potentially lowering R-Precision when token representation is weak, partly because our decoding is not text-guided. 
% However, with RVQs, a popular vector quantization on recent works [13,34], this issue is \textit{effectively mitigated}, shown by Tab.~\textcolor{LimeGreen}{2} and Tab.~\ref{tab:table_rebuttal_bamm}, where our R-precision remains on par with the original models.
\paragraph{Natural Motion Generation.} To evaluate DisCoRD’s effectiveness in decoding predicted tokens, we use models trained in stage 1 to assess their performance in decoding tokens generated by pretrained token predictors. As shown in Table~\ref{tab:quantitative_eval}, our method consistently outperforms baseline models, particularly in terms of FID,  achieving state-of-the-art performance in naturalness. We observe that for T2M-GPT, which employs a vanilla VQ-VAE with limited representational capacity, there is a slight decline in faithfulness, as a single token can map to multiple motions in DisCoRD. However, with RQVAEs \cite{RQVAE}, a popular quantization method on recent works \cite{pinyoanuntapong2024bamm, guo2024momask}, DisCoRD performs on par and even increase faithfullness, shown by R-Precision and MM-Dist. These results indicate that when paired with a decent tokenizer, DisCoRD can significantly boost naturalness without sacrificing faithfulness, highlighting its potential as a default decoder replacement for discrete motion generation models.

% We observed that as a single token can map to multiple motions in DisCoRD, for T2M-GPT which has limited representation power can lower R-precision 

% and demonstrating that an expressive decoder enhances not only reconstruction quality but also motion generation. We found that 

% While R-precision scores for T2M-GPT indicate a decrease in faithfulness, our results with MoMask and BAMM show comparable and sometimes even increases faithfulness. 

% paired with a powerful tokenizer, RQVAE—show that DisCoRD attains high naturalness without sacrificing faithfulness, highlighting its potential as a default decoder replacement for discrete methods.

\paragraph{Performance on Various Tasks.} To validate our approach as a general method for enhancing naturalness in discrete-based human motion generation, we train DisCoRD on co-speech gesture and music-driven dance generation, conducting a comparative analysis against baseline models. As shown in Table~\ref{tab:cospeech} and Table~\ref{tab:dance}, our method consistently outperforms baseline models across both tasks, achieving superior performance on sJPE and standard evaluation metrics. We present additional evaluation results in Supplementary Section D.
% ~\ref{sec:additional_quantitaive_results}.
% show superior performance, boosting original discrete methods especially in terms of FID. 

% For text-to-motion generation, we perform comparative analysis aginst state-of-the-art methods, presented in Table~\ref{tab:quantitative_eval}. The results demonstrate that our method consistently surpasses baseline models, particularly in FID, and achieves state-of-the-art performance when adapted to current leading approach, Momask. Each experiment is repeated 20 times, and the reported metric values represent the mean with a 95\% statistical confidence interval. In addition to existing metrics, we conduct a comparative analysis of sJPE in Table~\ref{tab:stage1}. We note that sJPE, like MPJPE, requires precise time alignment and is therefore measurable only in the reconstruction stage. Our results demonstrate that our method consistently outperforms baseline models in sJPE, highlighting its ability to produce more natural motion.

\begin{figure}[t] \centering
% \vspace{-0.3em}
    \includegraphics[width=0.48\textwidth]{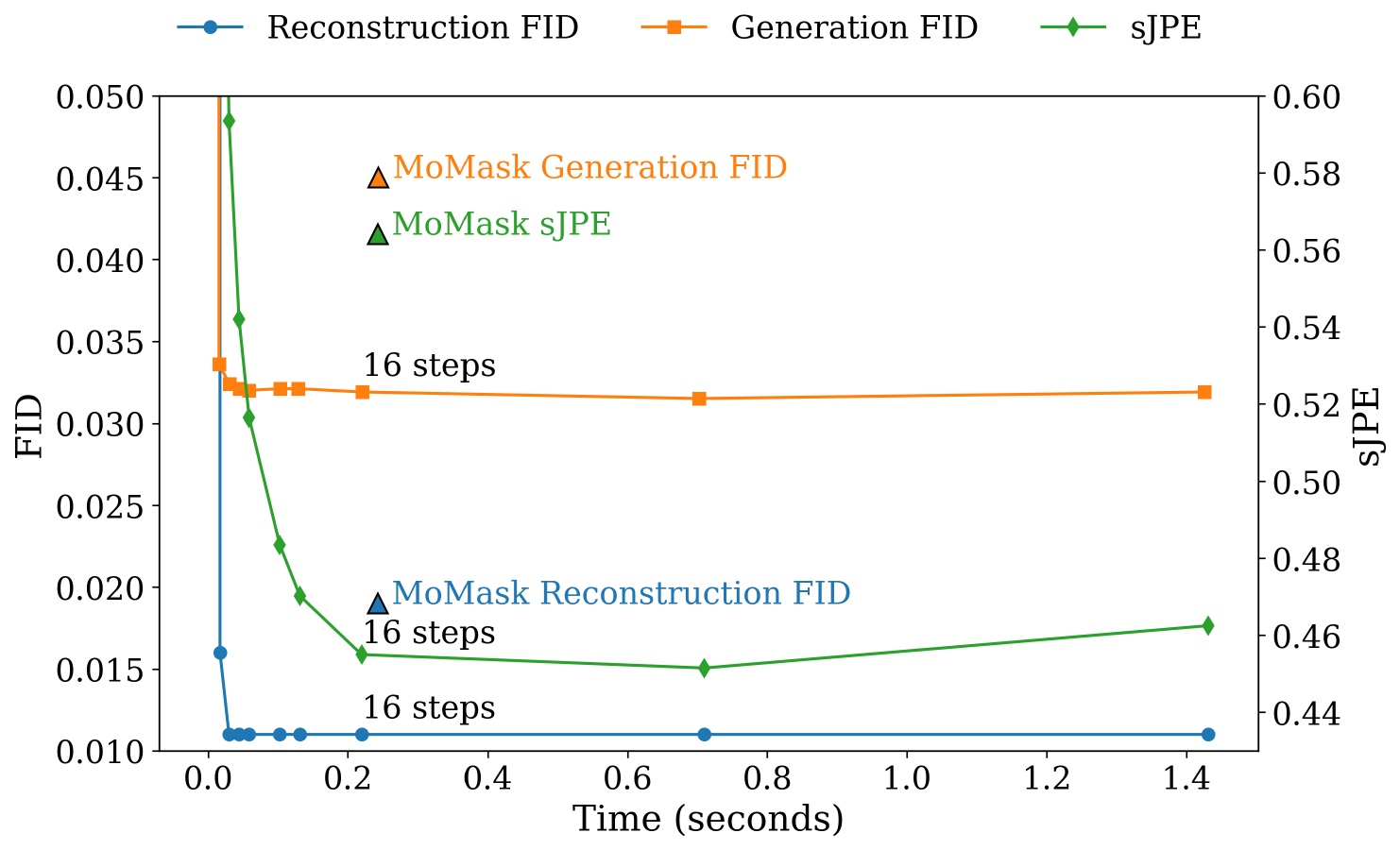}
    \caption{\textbf{Decoding efficiency comparison.} We report the average decoding time for a batch of 32 token sequences on an NVIDIA RTX 4090 Ti, averaged over 20 trials on the HumanML3D test set. DisCoRD achieves more better performance on motion naturalness at a comparable decoding speed to MoMask and can even decode significantly faster while maintaining superior performance.} \label{fig:time}
    \vspace{-0.2em}
\end{figure}

\paragraph{Effect of Sample Steps. }By selecting the rectified flow algorithm from among the various diffusion model variants, we exploit its efficient transport mechanism to achieve inference speeds comparable to baseline models. As shown in Figure~\ref{fig:time}, we evaluated the decoding times for both Momask and our model on tokens generated by a pretrained token generator. At the default setting of 16 sampling steps, our model achieves decoding speeds on par with MoMask while delivering superior sJPE, stage 1 FID, and stage 2 FID. Furthermore, by reducing the sampling steps, our method can decode tokens significantly fasterer than MoMask, maintaining comparable or enhanced FID and sJPE performance. Additionally, although it was not the primary focus of this experiment, we observed that sJPE responds more sensitively to changes in sampling steps compared to FID, further confirming that our sJPE metric effectively captures subtle variations in motion quality. Detailed decoding times are reported in the Supplementary Table E.

\paragraph{Effect of Residual Quantization.} We evaluate the impact of the representational capacity of motion tokens on DisCoRD by varying the residual quantization (RQ) levels in MoMask. As shown in Tab.~\ref{tab:rq}, while DisCoRD consistently improves FID across all RQ levels, its R-Precision gains over each MoMask baseline increase with higher RQ levels, suggesting that DisCoRD better leverages richer token representations. This result is consistent with Table~\ref{tab:quantitative_eval} where R-Precision drops for T2M-GPT (based on standard VQ-VAE) but maintains performance for MoMask and BAMM.

% DisCoRD builds on pretrained VQ-VAEs, and its performance depends on the capacity of their discretized representations. When this capacity is limited, a single token may correspond to multiple plausible motions, causing decoding ambiguity and reducing R-Precision. RQVAE addresses this limitation by enriching token representations through residual quantization, as implemented in MoMask and BAMM. 

% DisCoRD consistently improves FID across all RQ levels. Notably, its R-Precision gains over each MoMask baseline increase with higher RQ levels, indicating that DisCoRD more effectively exploits richer representations. A similar trend appears in Table~\ref{tab:quantitative_eval}: R-Precision drops for T2M-GPT (based on standard VQ-VAE) but improves for MoMask and BAMM.

\begin{table}[t]
    \centering
    % \vspace{-0.2em}
    \resizebox{\linewidth}{!}{
    \begin{tabular}{l c c c c c c}
    \toprule
    \multirow{2}{*}{Methods} & \multicolumn{1}{c}{Recon} & & \multicolumn{4}{c}{Generation}\\
    \cline{2-2} \cline{4-7}
    \addlinespace[0.2em]
        ~ & FID$\downarrow$ & & FID$\downarrow$ & RP@1$\uparrow$ & RP@2$\uparrow$ & RP@3$\uparrow$\\
    \midrule
    MoMask‑R0 & 0.125 && 0.082 & 0.505 & 0.696 & 0.793\\
    \rowcolor{blue!13}
   +Ours &0.061\makebox[0pt][l]{\textcolor{blue}{\scriptsize (+51\%)}}\hspace{2.5em} && 0.068\makebox[0pt][l]{\textcolor{blue}{\scriptsize (+17\%)}}\hspace{2.5em} & 0.491\makebox[0pt][l]{\textcolor{red}{\scriptsize (-2.8\%)}}\hspace{2.5em} & 0.682\makebox[0pt][l]{\textcolor{red}{\scriptsize (-2.0\%)}}\hspace{2.5em} &  0.782\makebox[0pt][l]{\textcolor{red}{\scriptsize (-1.4\%)}}\hspace{1.5em}\hphantom{00}\\
    MoMask‑R1 & 0.066 && 0.058 & 0.517 & 0.710 & 0.805\\
    \rowcolor{blue!13}
    +Ours &
        0.024\makebox[0pt][l]{\textcolor{blue}{\scriptsize (+64\%)}}\hspace{2.5em} &
        & 0.037\makebox[0pt][l]{\textcolor{blue}{\scriptsize (+36\%)}}\hspace{2.5em} &
        0.510\makebox[0pt][l]{\textcolor{red}{\scriptsize (-1.4\%)}}\hspace{2.5em} &
        0.703\makebox[0pt][l]{\textcolor{red}{\scriptsize (-1.0\%)}}\hspace{2.5em} &
        0.800\makebox[0pt][l]{\textcolor{red}{\scriptsize (-0.6\%)}}\hspace{1.5em}\hphantom{00}\\
    MoMask‑R2 & 0.046 && 0.053 & 0.520 & 0.712 & 0.806\\
    \rowcolor{blue!13}
    +Ours &
        0.018\makebox[0pt][l]{\textcolor{blue}{\scriptsize (+61\%)}}\hspace{2.5em} &
        & 0.031\makebox[0pt][l]{\textcolor{blue}{\scriptsize (+42\%)}}\hspace{2.5em} &
        0.517\makebox[0pt][l]{\textcolor{red}{\scriptsize (-0.6\%)}}\hspace{2.5em} &
        0.709\makebox[0pt][l]{\textcolor{red}{\scriptsize (-0.4\%)}}\hspace{2.5em} &
        0.806\makebox[0pt][l]{\textcolor{gray}{\scriptsize (0.0\%)}}\hspace{1.5em}\hphantom{}\\
    MoMask‑R4 & 0.024 && 0.047 & 0.519 & 0.711 & 0.807\\
    \rowcolor{blue!13}
    +Ours &
        0.011\makebox[0pt][l]{\textcolor{blue}{\scriptsize (+54\%)}}\hspace{2.5em} &
        & 0.032\makebox[0pt][l]{\textcolor{blue}{\scriptsize (+32\%)}}\hspace{2.5em} &
        0.520\makebox[0pt][l]{\textcolor{blue}{\scriptsize (+0.2\%)}}\hspace{2.7em} &
        0.711\makebox[0pt][l]{\textcolor{gray}{\scriptsize (0.0\%)}}\hspace{1.5em} &
        0.806\makebox[0pt][l]{\textcolor{red}{\scriptsize (-0.1\%)}}\hspace{1.5em}\hphantom{00}\\
    MoMask‑R5 & 0.019 && 0.045 & 0.521 & 0.713 & 0.807\\
    \rowcolor{blue!13}
    +Ours &
        0.011\makebox[0pt][l]{\textcolor{blue}{\scriptsize (+42\%)}}\hspace{2.5em} &
        & 0.032\makebox[0pt][l]{\textcolor{blue}{\scriptsize (+29\%)}}\hspace{2.5em} &
        0.524\makebox[0pt][l]{\textcolor{blue}{\scriptsize (+0.6\%)}}\hspace{2.6em} &
        0.715\makebox[0pt][l]{\textcolor{blue}{\scriptsize (+0.3\%)}}\hspace{2.7em} &
        0.809\makebox[0pt][l]{\textcolor{blue}{\scriptsize (+0.2\%)}}\hspace{1.5em}\hphantom{00}\\
    \bottomrule
    \end{tabular}
    }
    % \vspace{-0.4em}
    \caption{\textbf{Effect of residual quantization levels (R)}. in MoMask on DisCoRD performance. Higher R yields greater R-Precision (RP) gains, indicating better use of richer representations.}
    \vspace{-0.5em}
    \label{tab:rq}
\end{table}

\paragraph{Ablation Studies. }
% To evaluate the contribution of each component to model performance, 
We conduct ablation studies on DisCoRD’s each component shown in 
Table~\ref{tab:ablation}.
First, we compare our decoding strategy with post refinement methods which refine output motions from MoMask's decoder. Feedforward convolution layers show little improvement and rectified flow post refinement falls short in all metrics compared to ours.
Second, we examine alternative projection mechanisms. While up-convolution or repeat followed by a linear layer show strong reconstruction performance, they fail to decode natural motion from generated token sequences (unseen at training) shown by low FID. 
Additionally, incorporating attention into the U-Net backbone and using full motion sequences instead of windowed motion segments result in performace degradation. This indicates that focusing on localized motion segments enhances the model’s generalization capability, particularly in stage 2.
% presents the results on the HumanML3D test set, assessing the impact of each component on reconstruction and generation quality. 
% This is because  as they make conditions on a sequence level, while ours operate on each tokens.
% our condition projection layer can decode unseen (generated) token sequences 
% resulted in significant performance degradation, underscoring the effectiveness of our approach. 
\begin{table}[t]
    \centering
    \resizebox{\linewidth}{!}{
    \begin{tabular}{l c c c c c}
    \toprule
    % \multirow{2}{*}{Methods} & \multicolumn{4}{c}{\DN (Coarse-grained)} & & \multicolumn{4}{c}{HumanAct(Fine-grained)} \\
    % \cline{2-5}
    % \cline{7-10}
    %                 & FID$\downarrow$ & Accuracy$\uparrow$ & Diversity$\rightarrow$& MModality$\rightarrow$ &  & FID$\downarrow$ & Accuracy$\uparrow$ & Diversity$\rightarrow$ & MModality$\rightarrow$\\
    \multirow{2}{*}{Methods}  & \multicolumn{2}{c}{Reconstruction} & & \multicolumn{2}{c}{Generation}\\

    \cline{2-3}
    \cline{5-6}
    \addlinespace[0.2em] 
       ~ & FID$\downarrow$ & sJPE$\downarrow$ &  & FID$\downarrow$ & MM-Dist$\downarrow$  \\
    \hline
    \midrule
    MoMask & \et{0.019}{.001} & 0.512 && \et{0.051}{.002} & \et{2.957}{.008} \\
    + Post Refinement (FF model) & \et{0.028}{.000}  & 0.481 & & \et{0.044}{.002} & \et{2.962}{.006}  \\
    + Post Refinement (RF model) & \et{0.013}{.000}  & 0.489  & & \et{0.035}{.002} & \et{2.955}{.008} \\
    \textbf{+ DisCoRD (Ours)}& \et{0.011}{.000} & 0.385 && \etb{0.032}{.002} & \etb{2.938}{.010} \\

    \hline
    % \rowcolor{lightgray}
    % \multicolumn{6}{c}{\textit{\textbf{Ablations}}} \\
    \hline
        \addlinespace[0.2em] 
        Ours (Upconv) & \et{0.010}{.000}  & 0.375 & & \et{0.039}{.003} & \et{2.943}{.006} \\
        Ours (Repeat \& Linear) & \et{0.011}{.001}  & \textbf{0.342} & & \et{0.038}{.001} & \et{2.947}{.008} \\
        Ours (w/ Attention) & \et{0.020}{.000}  & 0.384 & & \et{0.043}{.002} & \et{2.983}{.009} \\
        Ours (w/ Full Motion Sequence)

        % Windowed Motion)
        & \etb{0.008}{.000}  & 0.385 & & \et{0.038}{.002} & \et{2.952}{.009} \\
        % Ours (DiT backbone) & \etb{9.999}{.001} & \textbf{29.5} && \etb{0.051}{.002} & \etb{2.957}{.008} \\
    \bottomrule
    \end{tabular}
    }
\caption{\textbf{Ablation studies.} Evaluation on the HumanML3D test set assessing the impact of decoding strategies, projection methods, and training strategies. FF and RF denote feedforward and rectified flow model, respectively.}

    % We perform ablations on the projection, attention, and motion windowing strategies during training on the HumanML3D dataset. 
    % Refinement methods indicate post refinement on decoded raw motions. 
    \label{tab:ablation}
    \vspace{-1.3em}
\end{table}

\subsection{Qualitative Results}
% To validate the superiority of DisCoRD in generating both faithful and natural human motions, 
% \paragraph{User Studies.} We conduct two user studies to (1) validate our motivation and method effectiveness and (2) evaluate how well sJPE aligns with human perception. The first study, shown in Figure~\ref{fig:user_study}, indicates that the discrete model Momask outperforms the continuous model MDM in faithfulness but lags in naturalness. In contrast, DisCoRD surpasses both, demonstrating its ability to generate motion that is both natural and faithful. In the second study, we find that sJPE exhibits 2.7 times higher correlation with human preference for naturalness compared to MPJPE, highlighting its effectiveness in evaluating sample-wise motion naturalness. Details of user studies are in Supplementary Section C.4 and E.2.

% Additional qualitative comparisons and details on user studies are provided in the Supplementary Section C and E.
% \begin{figure}[t] \centering
%     \includegraphics[width=0.48\textwidth]{figures/qual.pdf}
%     \caption{\textbf{Qualitative comparisons} on the test set of HumanML3D.} \label{fig:qualitative}
%     % \vspace{-1.3em}
% \end{figure}

We visualize motion trajectories in Figure~\ref{fig:sjpe_comparison}, where DisCoRD, unlike discrete methods, produces smooth and expressive motion with low sJPE. 
% Additionally, as shown in Figure~\ref{fig:qualitative}, our method generates motions samples that closely align with textual descriptions while preserving a high degree of naturalness. 
Extensive visualizations are provided in Supplementary Sections C.3 and E.1 with supporting user studies in Sections C.4 and E.2.
% that continuous models are generally more effective at generating natural motions but often struggle to adhere strictly to conditions, whereas discrete models tend to excel in condition faithfulness but often lack naturalness of motion. 
% demonstrates its effectiveness by surpassing both discrete and continuous models in terms of both naturalness and faithfulness, achieving a balanced and superior performance. 
% ~\ref{sec:additional_visualizations}.

%% file: sec/5_Conclusion.tex
\section{Conclusion and Discussion}
\label{sec:Conclusion}
In this paper, we present DisCoRD, a novel approach that decodes discrete motion tokens to natural, dynamic human motion using rectified flow.
% to human motion generation 
% that effectively combines the naturalness of continuous representations with the faithfulness of discrete quantization methods. 
To demonstrate gains in naturalness, we also introduce symmetric Jerk Percentage Error (sJPE), specifically designed to capture subtle artifacts that were overlooked by traditional metrics. Extensive experiments across text-to-motion, co-speech gesture, and music-to-dance tasks demonstrate that DisCoRD consistently achieves state-of-the-art performance, providing a versatile solution adaptable to various discrete-based motion generation frameworks. While we experimented only on human motion generation, a promising direction would be to expand our framework to discrete talking face, hand motion, or even whole body motion generation.

% \paragraph{Limitations.}
% Since our method fundamentally builds on baseline models, it inherits certain limitations associated with these models. For instance, if the baseline model cannot generate variable motion lengths, our approach, which shares the same token generation model, may also be restricted by this limitation. Additionally, our current method requires a pretrained model as a starting point. However, our framework is inherently designed to support end-to-end training, allowing for the possibility of constructing an entirely new encoder from scratch. We leave this potential expansion as an avenue for future work.

% \paragraph{Acknowledgments.} This study was supported by Institute of Information & communications Technology Promotion(IITP) grant funded by the Korea government(MSIT) (No. RS-2024-00457882 Artificial Intelligence Research Hub Project). This work was supported by Culture, Sports and Tourism R\&D Program through the Korea Creative Content Agency grant funded by the Ministry of Culture, Sports and Tourism in 2024 (Project Name:Development of multimodal UX evaluation platform technology for XR spatial responsive content optimization, Project Number: RS-2024-00361757). T.-H. Oh was partially supported by Institute of Information & Communications Technology Planning & Evaluation (IITP) grant funded by the Korea government(MSIT) (No.RS-2023-00225630, Development of Artificial Intelligence for Text-based 3D Movie Generation). 
\paragraph{Acknowledgments.} 
This study was supported by the Institute of Information \& Communications Technology Planning \& Evaluation (IITP) grant funded by the Korea government (MSIT) (No. RS-2024-00457882, Artificial Intelligence Research Hub Project). This work was also supported by the Culture, Sports and Tourism R\&D Program through the Korea Creative Content Agency, funded by the Ministry of Culture, Sports and Tourism in 2024 (Project Name: Development of multimodal UX evaluation platform technology for XR spatial responsive content optimization, Project Number: RS-2024-00361757). T.-H. Oh was partially supported by the IITP grant funded by the Korea government (MSIT) (No. RS-2023-00225630, Development of Artificial Intelligence for Text-based 3D Movie Generation).

%% file: sec/X_suppl.tex
\clearpage
\setcounter{section}{0}
\setcounter{table}{0}
\setcounter{figure}{0}
\renewcommand{\thesection}{\Alph{section}} 
\renewcommand{\thesubsection}{\Alph{section}.\arabic{subsection}}
\renewcommand{\thefigure}{\Alph{figure}}
\setcounter{table}{0}
\renewcommand{\thetable}{\Alph{table}}

\maketitlesupplementary
\appendix

This supplementary material is organized as follows: Section~\ref{sec:implementation_details} details the implementation of DisCoRD. Section~\ref{sec:datasets_and_evaluations} provides additional information on the datasets and evaluation metrics. Section~\ref{sec:additional_analysis_on_sJPE} offers a comprehensive analysis of sJPE. Section~\ref{sec:additional_quantitaive_results} presents quantitative results excluded from the main paper. Section~\ref{sec:additional_visualizations} includes additional qualitative results. \textit{\textbf{We highly recommend viewing the accompanying video}}, as static images are insufficient to fully convey the intricacies of motion.

\section{Implementation Details}
\label{sec:implementation_details}
Table \ref{tab:implementation} provides an overview of the implementation details for our method. These configurations were employed to train the DisCoRD decoder using the pretrained Momask~\cite{guo2024momask} quantizer. Specifically, the 512-dimensional codebook embeddings from MoMask are projected into the conditional channel dimension. This projection is concatenated with Gaussian noise of the same dimensionality as the output channel. The concatenated representation is subsequently projected into the input channel dimension of the U-Net architecture. The U-Net processes this input and transforms it back into the output channel dimension, generating the final output shape. For training, we used an input window size of 64 and trained the model for 35 hours on a single NVIDIA RTX 4090 Ti GPU.

\begin{table}[h!]
\centering
\scalebox{1.0}{
    \begin{tabular}{ll|l}
    \toprule
    \multicolumn{3}{c}{\textbf{Training Details}} \\
    \midrule
    Optimizer        & & AdamW $(0.9, 0.999)$ \\
    LR               & & 0.0005\\
    LR Decay Ratio   & & 0 \\
    LR Scheduler     & & Cosine \\
    Warmup Epochs    & & 20\\
    Gradient Clipping& & 1.0 \\
    Weight EMA              & & 0.999\\
    Flow Loss        & & MSE Loss \\
    \midrule
    Batch Size       & & 768\\
    Window Size      & & 64\\
    Steps            & & 481896 \\
    Epochs           & & 200\\
    \toprule
    \multicolumn{3}{c}{\textbf{Model Details}} \\
    \midrule
    Input Channels   & & 512 \\
    Output Channels  & & 263\\
    Condition Channels & & 256 \\
    Activation       & & SiLU \\
    Dropout          & & 0\\
    Width            & & (512, 1024) \\
    \# Resnet / Block & & 2 \\ 
    \# Params        & & 66.9M \\
    \end{tabular}
}
\caption{\textbf{Implementation details} for training the DisCoRD decoder on the HumanML3D dataset using the pretrained Momask quantizer.}
\label{tab:implementation}
\end{table}

\section{Datasets and Evaluations}
\label{sec:datasets_and_evaluations}
In this section, we provide additional explanations regarding the co-speech gesture generation and music-to-dance generation tasks that we were unable to describe in detail in the main paper.

% \paragraph{Datasets. } 
\subsection{Datasets. }For the co-speech gesture generation task, we utilized the SHOW dataset~\cite{Talkshow}, a 3D holistic body dataset comprising 26.9 hours of in-the-wild talking videos. For the music-driven dance generation task, we used a mixed dataset combining AIST++~\cite{aist++} and HumanML3D~\cite{guo_generating_3dhuman} where AIST++ is a large-scale 3D dance dataset created from multi-camera videos accompanied by music of varying styles and tempos, containing 992 high-quality pose sequences in the SMPL format.

% \paragraph{Evaluations. } 
\subsection{Evaluations. } To evaluate the co-speech gesture generation task, we used Frechet Gesture Distance (FGD)~\cite{FGD}, which measures the difference between the latent distributions of generated and real motions. Since our focus is on body movements, we reported body FGD, which quantifies differences specifically for the body part, for ProbTalk~\cite{probtalk}. For TalkSHOW~\cite{Talkshow}, which only utilizes holistic FGD—a metric that measures differences across the entire motion, including the face and hands—we reported the holistic FGD. To evaluate the music-to-dance generation task, we utilized $\mathrm{Dist_k}$, which quantifies the distributional spread of generated dances based on kinetic features, and $\mathrm{Dist_g}$, which does the same for geometric features, as proposed in~\cite{EDGE_fidcannotcatch}. A smaller difference between the distributions of the generated motion and the ground truth motion indicates that the $\mathrm{Dist_k}$ and $\mathrm{Dist_g}$ values of the generated motion align closely with those of the ground truth, reflecting a similar level of distributional spread.

\section{Additional Analysis on sJPE}
\label{sec:additional_analysis_on_sJPE}
To evaluate the sample-wise naturalness of reconstructed motions, we introduce the symmetric Jerk Percentage Error (sJPE), as defined in Equation~\ref{eq:mds} of the main paper. We present detailed formulations of \textit{Noise sJPE} and \textit{Static sJPE}, supported by analysis using generated motion samples. Furthermore, qualitative comparisons highlight the effectiveness of DisCoRD against state-of-the-art discrete methods. Finally, we investigate the alignment of sJPE with human preference to validate its perceptual relevance.

\subsection{Visualization of Fine-Grained Motion}
To analyze fine-grained motion trajectories, we follow a three-step procedure. First, we select a joint for visualization, typically hand joints due to their high dynamism, and track their positional changes over time. Second, we apply a Gaussian filter to smooth the trajectory, reducing noise. Finally, we compute the difference between the smoothed and original trajectories to isolate fine-grained motion components. This method allows for detailed evaluation of frame-wise noise and under-reconstructed regions in motion trajectories. The visualizations in Figure 5 of the main paper and the qualitative samples in the supplementary material are generated using this process.

\subsection{Details on sJPE. } 
Within the symmetric Jerk Percentage Error (sJPE), we define two components: Noise sJPE and Static sJPE. These isolate the instances where the predicted jerk overestimates or underestimates the ground truth jerk, respectively.

\paragraph{Noise sJPE and Static sJPE.} \textit{Noise sJPE} captures the average overestimation of jerk in the predicted motion signal, meaning frame-wise noise, corresponding to cases where \( J_{\text{pred},t} > J_{\text{true},t} \). It is defined as:
\begin{equation}
\text{Noise sJPE} = \frac{1}{n} \sum_{t=1}^{n} \frac{\max\left(0, J_{\text{pred},t} - J_{\text{true},t}\right)}{|J_{\text{true},t}| + |J_{\text{pred},t}|}.
\label{eq:noise_sjpe}
\end{equation}
The operator \( \max(0, x) \) ensures that only positive differences contribute to Noise sJPE, separating overestimations from underestimations.

\begin{figure}[h] \centering
    \includegraphics[width=0.48\textwidth]{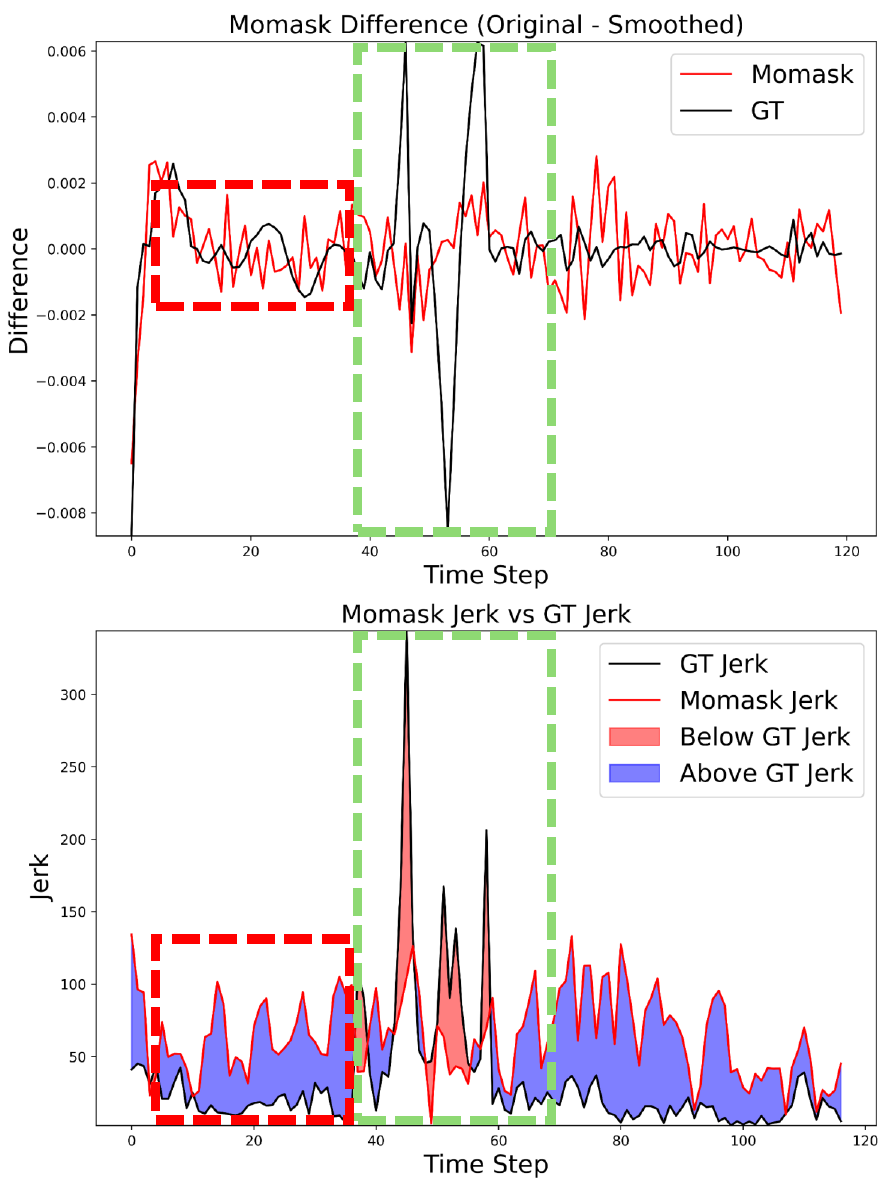}
    \caption{\textbf{Relationship between fine-grained trajectory and jerk:} Frame-wise noise in predicted motions, highlighted in the red box, results in higher jerk values compared to the ground truth, represented by the blue areas. The sum of the blue areas corresponds to Noise sJPE. Conversely, under-reconstruction in predicted motions, highlighted in the green box, leads to lower jerk values compared to the ground truth, represented by the red areas. The sum of the red areas corresponds to Static sJPE.} \label{fig:jerk_area_explanation}
\end{figure}

Noise sJPE can be seen on the red box of Figure~\ref{fig:jerk_area_explanation}. Time steps where motion trajectory is noisy compared to ground truth motion show bigger jerk. The area under the predicted jerk and above the ground truth jerk, shown in blue area, is proportional to the Noise sJPE, meaning frame-wise noise.

\textit{Static sJPE} measures the average underestimation of jerk, meaning lack of dynamism in the predicted motion, corresponding to cases where \( J_{\text{pred},t} \leq J_{\text{true},t} \). It is defined as:
\begin{equation}
\text{Static sJPE} = \frac{1}{n} \sum_{t=1}^{n} \frac{\max\left(0, J_{\text{true},t} - J_{\text{pred},t}\right)}{|J_{\text{true},t}| + |J_{\text{pred},t}|}.
\label{eq:static_sjpe}
\end{equation}
Static sJPE can be seen on the green box of Figure~\ref{fig:jerk_area_explanation}. Time steps where motion trajectory is under-reconstructed compared to ground truth motion show smaller jerk. The area above the predicted jerk and under the ground truth jerk, shown in red area, is proportional to the Static sJPE, meaning under reconstructed motions. 

The overall sJPE can be expressed as the sum of \textit{Noise sJPE} and \textit{Static sJPE}:
\begin{equation}
\text{sJPE} = \text{Noise sJPE} + \text{Static sJPE}.
\end{equation}

These formulations provide a measure of prediction accuracy by separately accounting for the tendencies of the predictive model to overestimate or underestimate the true motion jerks.

\begin{figure*}[ht]
    \centering
    \includegraphics[width=0.99\linewidth]{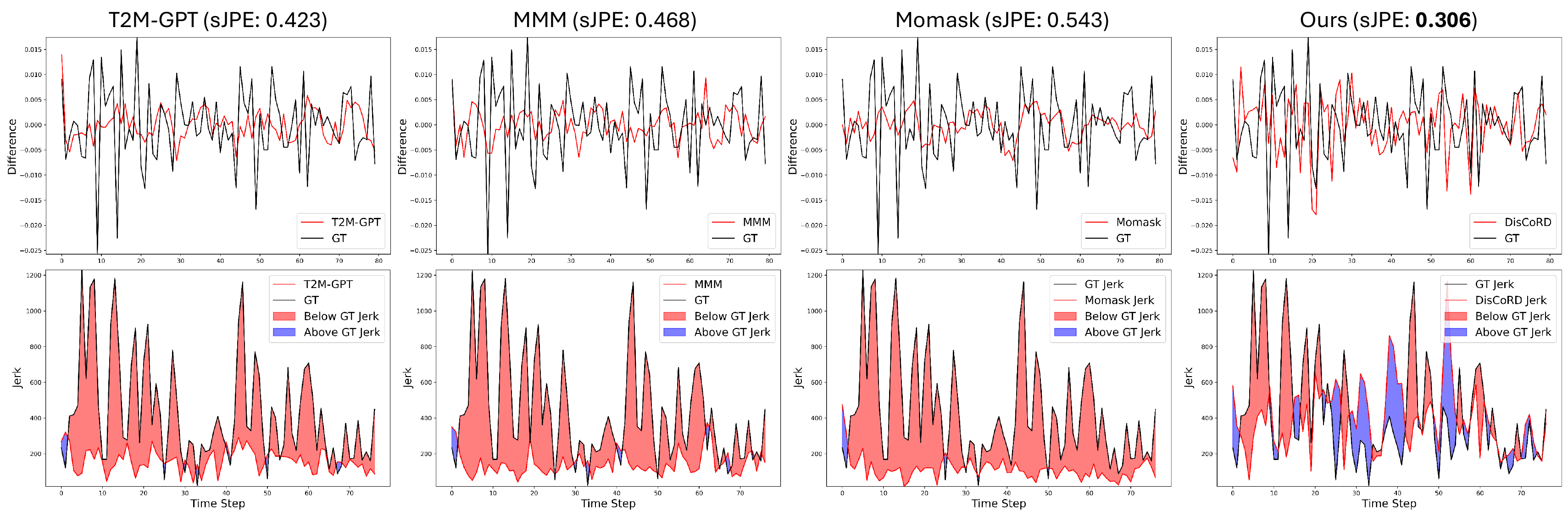}
    \caption{\textbf{Joint Trajectory and Jerk: Under-Reconstruction in Discrete Methods}  
DisCoRD effectively reduces the red area, demonstrating its capability to reconstruct dynamic motion accurately. This improvement is also reflected in the lower sJPE value.}
    \label{fig:big_qaulitative_chart_supple_underrecon}
\end{figure*}
\begin{figure*}[ht]
    \centering
    \includegraphics[width=0.99\linewidth]{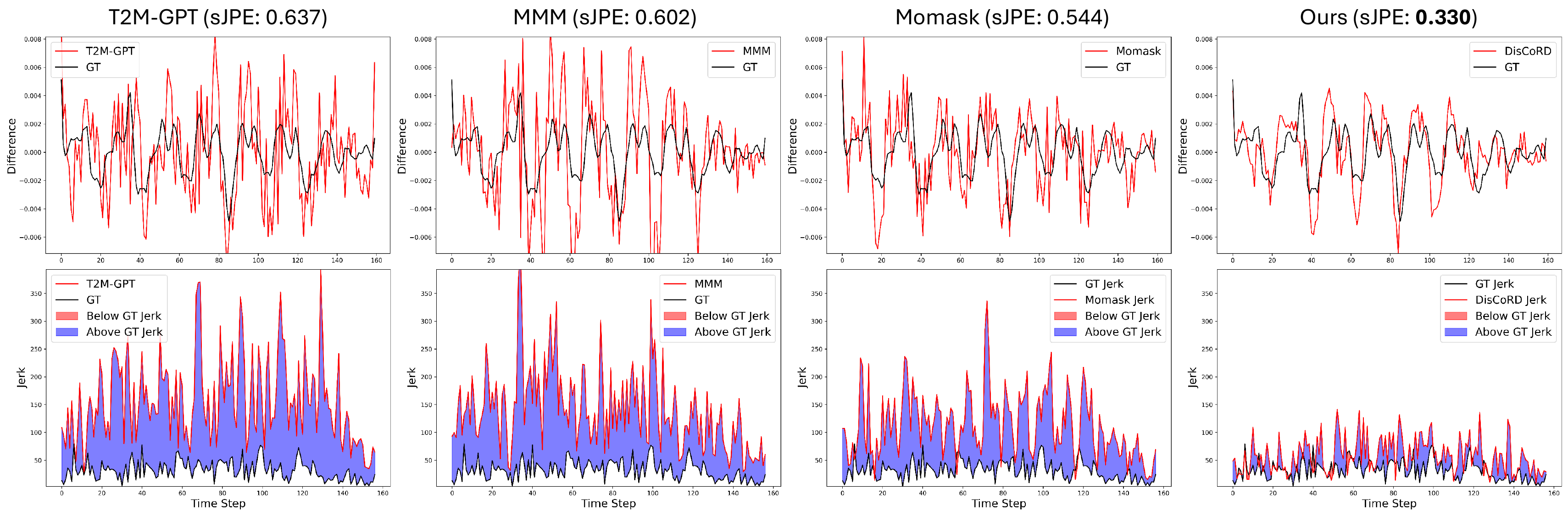}
    \caption{
\textbf{Joint Trajectory and Jerk: Frame-Wise Noise in Discrete Methods}
DisCoRD significantly reduces the blue area, indicating its ability to generate smooth motions that closely resemble the ground truth. This improvement is further reflected in the lower sJPE value. }
    \label{fig:big_qaulitative_chart_supple_noise}
\end{figure*}

\subsection{Qualitative Results on Joint Trajectory and Jerk}  
We present a series of figures demonstrating the effectiveness of DisCoRD in reconstructing smooth and dynamic motion. For each sample, the first row visualizes the motion trajectory, while the second row plots the corresponding jerk at each time step, with the calculated sJPE displayed at the top. This visualization enables detailed analysis of fine-grained trajectories in predicted motions and highlights the contributions of Noise sJPE and Static sJPE to the overall sJPE.  

\begin{figure*}[ht]
    \centering
    \includegraphics[width=0.99\linewidth]{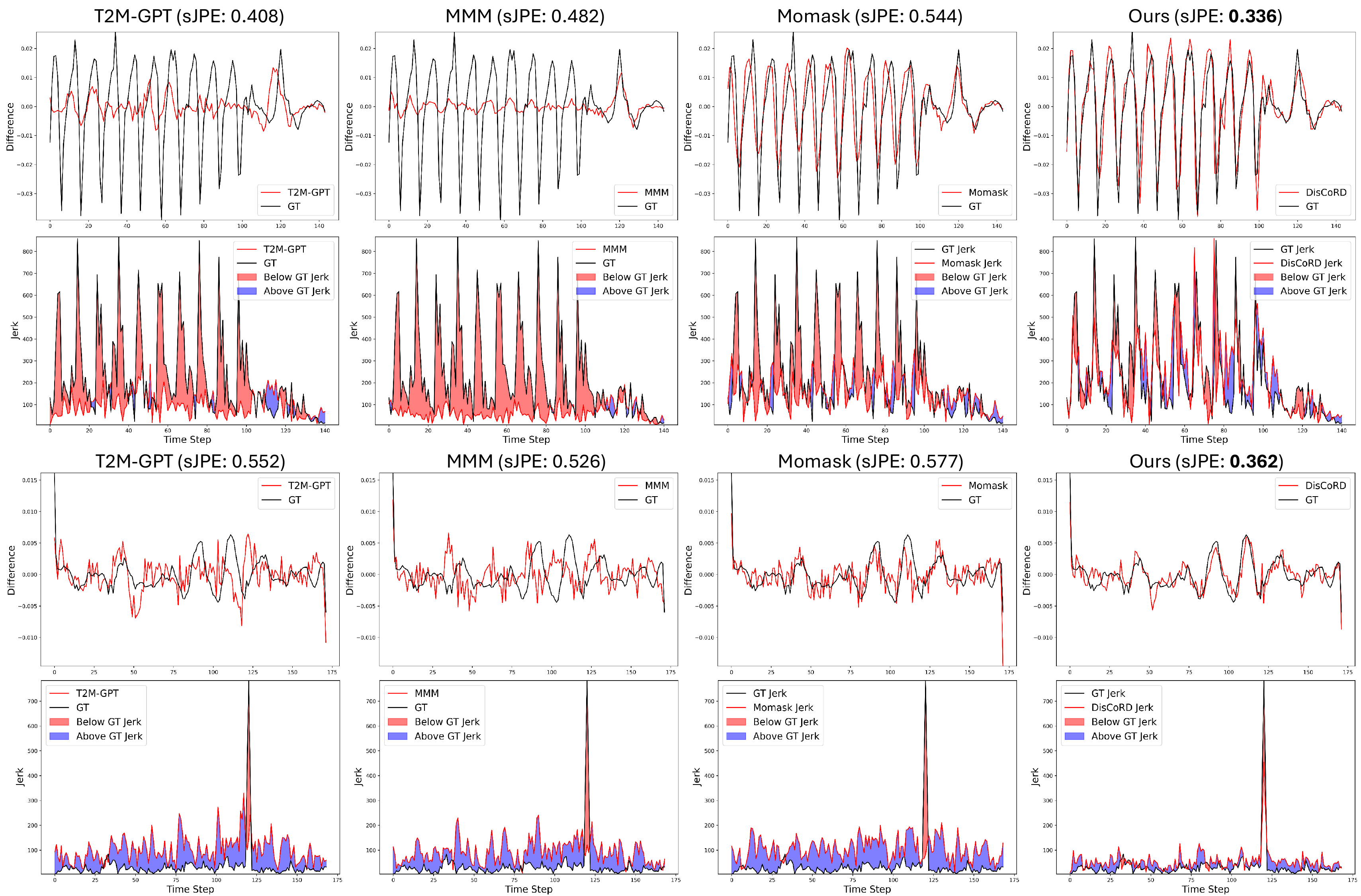}
    \caption{\textbf{Joint Trajectory and Jerk: Both Frame-Wise Noise and Under-Reconstruction in Discrete Methods}
DisCoRD addresses both frame-wise noise and under-reconstruction by simultaneously reducing the blue and red areas. This demonstrates its ability to generate smooth and dynamic motions, closely aligning with the ground truth. This further supported by the lower sJPE values.}
    \label{fig:big_qaulitative_chart_supple_both}
\end{figure*}

We compare DisCoRD with recent discrete methods, including T2M-GPT~\cite{zhang2023t2m}, MMM~\cite{pinyoanuntapong2024mmm}, and Momask~\cite{guo2024momask}. Motion samples that illustrate under-reconstruction in discrete methods are presented in Figure~\ref{fig:big_qaulitative_chart_supple_underrecon}, while those that exhibit frame-wise noise are shown in Figure~\ref{fig:big_qaulitative_chart_supple_noise}. Samples showing both issues in discrete models are displayed in Figure~\ref{fig:big_qaulitative_chart_supple_both}. DisCoRD effectively reduces frame-wise noise while accurately reconstructing dynamic, fast-paced motions. This is shown in both the visualizations and the sJPE results.

\subsection{Correlation between sJPE and human perception. }To further verify that sJPE aligns with human judgment of naturalness, we conducted an additional user study. We asked participants to rank three models—MLD, MoMask, and ours—in order of naturalness, guided as Figure~\ref{fig:survey_2}. The user interface for this user study is shown in  Figure~\ref{fig:survey_2_q}. The rankings were scored such that the first place received 1 point, the second place 2 points, and the third place 3 points. Using these human scores, we calculated Pearson's correlation between the human scores and two metrics—MPJPE and sJPE— for each sample. During this process, we excluded the lowest 10\% of samples in terms of human score standard deviation among models, as these were considered indistinguishable by human evaluators. Our analysis revealed that the average Pearson's correlation between MPJPE and human scores was 0.181, whereas the correlation between sJPE and human scores was significantly higher at 0.483. This result demonstrates the effectiveness of sJPE in evaluating sample-wise naturalness.

% \begin{figure*}[t]
%     \centering
%     \includegraphics[width=0.99\linewidth]{figures/Paperswithcode.png}
% \caption{\textbf{Ranking on PapersWithCode.} DisCoRD outperform recent methods and matches concurrent approaches, only by training new decoders on pretrained models.}

%     \label{fig:paperswithcode}
% \end{figure*}

\section{Additional Quantitative Results}
\label{sec:additional_quantitaive_results}
\subsection{Performance on Text-to-Motion Generation. }
In Table~\ref{tab:quantitative_supp}, we present a comparison of our method against additional results from various text-to-motion models. Our method consistently achieves strong performance on the HumanML3D and KIT-ML~\cite{kit} test sets, even when evaluated alongside these additional models. While ReMoDiffuse achieves particularly strong performance on KIT-ML, it is worth noting that its performance benefits from the use of a specialized database for high-quality motion generation, which makes direct comparisons less appropriate.

\begin{table*}[t]
    \centering
    \scalebox{0.7}{
    \begin{tabular}{l l c c c c c c}
    \toprule
    \multirow{2}{*}{Datasets} & \multirow{2}{*}{Methods}  & \multicolumn{3}{c}{R Precision $\uparrow$} & \multirow{2}{*}{FID $\downarrow$} & \multirow{2}{*}{MultiModal Dist $\downarrow$} & \multirow{2}{*}{MultiModality $\uparrow$} \\
    \cline{3-5}
    \addlinespace[0.2em]  
    & & Top 1 & Top 2 & Top 3 \\
    \midrule
    \multirow{16}{*}{\makecell[c]{Human\\ML3D}} 
        & MDM~\cite{tevet2023human} & - & - & \et{0.611}{.007} & \et{0.544}{.044} & \et{5.566}{.027} & \ets{2.799}{.072} \\
        & MLD~\cite{chen2023executing} & \et{0.481}{.003} & \et{0.673}{.003} & \et{0.772}{.002} & \et{0.473}{.013} & \et{3.196}{.010} & \et{2.413}{.079} \\
        & MotionDiffuse~\cite{zhang2022motiondiffuse} & \et{0.491}{.001} & \et{0.681}{.001} & \et{0.782}{.001} & \et{0.630}{.001} & \et{3.113}{.001} & \et{1.553}{.042} \\
        & ReMoDiffuse~\cite{zhang2023remodiffuse} & \et{0.510}{.005} & \et{0.698}{.006} & \et{0.795}{.004} & \et{0.103}{.004} & \et{2.974}{.016} & \et{1.795}{.043} \\
        & Fg-T2M~\cite{wang2023fg} & \et{0.492}{.002} & \et{0.683}{.003} & \et{0.783}{.024} & \et{0.243}{.019} & \et{3.109}{.007} & \et{1.614}{.049} \\
        & M2DM~\cite{m2dm} & \et{0.497}{.003} & \et{0.682}{.002} & \et{0.763}{.003} & \et{0.352}{.005} & \et{3.134}{.010} & \etb{3.587}{.072} \\
        & M2D2M~\cite{m2d2m} & - & - & \et{0.799}{.002} & \et{0.087}{.004} & \et{3.018}{.010} & \et{2.115}{.079} \\
        & MotionGPT~\cite{zhang2024motiongpt} & \et{0.364}{.005} & \et{0.533}{.003} & \et{0.629}{.004} & \et{0.805}{.002} & \et{3.914}{.013} & \et{2.473}{.041} \\
        & MotionLLM~\cite{motionllm} & \et{0.482}{.004} & \et{0.672}{.003} & \et{0.770}{.002} & \et{0.491}{.019} & \et{3.138}{.010} & - \\
        & MotionGPT-2~\cite{motiongpt2} & \et{0.496}{.002} & \et{0.691}{.003} & \et{0.782}{.004} & \et{0.191}{.004} & \et{3.080}{.013} & \et{2.137}{.022} \\
        & AttT2M~\cite{zhong2023attt2m} & \et{0.499}{.003} & \et{0.690}{.002} & \et{0.786}{.002} & \et{0.112}{.006} & \et{3.038}{.007} & \et{2.452}{.051} \\
        & MMM~\cite{pinyoanuntapong2024mmm} & \et{0.504}{.003} & \et{0.696}{.003} & \et{0.794}{.002} & \et{0.080}{.003} & \et{2.998}{.007} & \et{1.164}{.041} \\
    \cline{2-8}
        \addlinespace[0.3em]  
        & T2M-GPT~\cite{zhang2023t2m} & \et{0.491}{.003} & \et{0.680}{.003} & \et{0.775}{.002} & \et{0.116}{.004} & \et{3.118}{.011} & \et{1.856}{.011} \\
        & \textbf{+ DisCoRD (Ours)} & \et{0.476}{.008} & \et{0.663}{.006} & \et{0.760}{.007} & \et{0.095}{.011} & \et{3.121}{.009} & \et{1.831}{.048} \\
    \cline{2-8}
        \addlinespace[0.3em]  
        & BAMM~\cite{pinyoanuntapong2024bamm} & \etb{0.525}{.002} & \etb{0.720}{.003} & \etb{0.814}{.003} & \et{0.055}{.002} & \etb{2.919}{.008} & \et{1.687}{.051} \\
        & \textbf{+ DisCoRD (Ours)} & \et{0.522}{.003} & \ets{0.715}{.005} & \ets{0.811}{.004} & \ets{0.041}{.002} & \ets{2.921}{.015} & \et{1.772}{.067} \\
    \cline{2-8}
        \addlinespace[0.3em]  
        & MoMask~\cite{guo2024momask} & \et{0.521}{.002} & \et{0.713}{.002} & \et{0.807}{.002} & \et{0.045}{.002} & \et{2.958}{.008} & \et{1.241}{.040} \\
        & \textbf{+ DisCoRD (Ours)} & \ets{0.524}{.003} & \ets{0.715}{.003} & \et{0.809}{.002} & \etb{0.032}{.002} & \et{2.938}{.010} & \et{1.288}{.043} \\
    \hline
    \midrule
    \multirow{16}{*}{\makecell[c]{KIT-\\ML}} 
        & MDM~\cite{tevet2023human} & - & - & \et{0.396}{.004} & \et{0.497}{.021} & \et{9.191}{.022} & \et{1.907}{.214} \\
        & MLD~\cite{chen2023executing} & \et{0.390}{.008} & \et{0.609}{.008} & \et{0.734}{.007} & \et{0.404}{.027} & \et{3.204}{.027} & \et{2.192}{.071} \\
        & MotionDiffuse~\cite{zhang2022motiondiffuse} & \et{0.417}{.004} & \et{0.621}{.004} & \et{0.739}{.004} & \et{1.954}{.062} & \et{2.958}{.005} & \et{0.730}{.013} \\
        & ReMoDiffuse~\cite{zhang2023remodiffuse} & \et{0.427}{.014} & \et{0.641}{.004} & \et{0.765}{.055} & \etb{0.155}{.006} & \et{2.814}{.012} & \et{1.239}{.028} \\
        & Fg-T2M~\cite{wang2023fg} & \et{0.418}{.005} & \et{0.626}{.004} & \et{0.745}{.004} & \et{0.571}{.047} & \et{3.114}{.015} & \et{1.019}{.029} \\
        & M2DM~\cite{m2dm} & \et{0.416}{.004} & \et{0.628}{.004} & \et{0.743}{.004} & \et{0.515}{.029} & \et{3.015}{.017} & \etb{3.325}{.037} \\
        & M2D2M~\cite{m2d2m} & - & - & \et{0.753}{.006} & \et{0.378}{.023} & \et{3.012}{.021} & \et{2.061}{.067} \\
        & MotionGPT~\cite{zhang2024motiongpt} & \et{0.340}{.002} & \et{0.570}{.003} & \et{0.660}{.004} & \et{0.868}{.032} & \et{3.721}{.018} & \et{2.296}{.022} \\
        & MotionLLM~\cite{motionllm} & \et{0.409}{.006} & \et{0.624}{.007} & \et{0.750}{.005} & \et{0.781}{.026} & \et{2.982}{.022} & - \\
        & MotionGPT-2~\cite{motiongpt2} & \et{0.427}{.003} & \et{0.627}{.002} & \et{0.764}{.003} & \et{0.614}{.005} & \et{3.164}{.013} & \ets{2.357}{.022} \\
        & AttT2M~\cite{zhong2023attt2m} & \et{0.413}{.006} & \et{0.632}{.006} & \et{0.751}{.006} & \et{0.870}{.039} & \et{3.039}{.021} & \et{2.281}{.047} \\
        & MMM~\cite{pinyoanuntapong2024mmm} & \et{0.404}{.005} & \et{0.621}{.006} & \et{0.744}{.005} & \et{0.316}{.019} & \et{2.977}{.019} & \et{1.232}{.026} \\
    \cline{2-8}
        \addlinespace[0.3em]  
        & T2M-GPT~\cite{zhang2023t2m} & \et{0.398}{.007} & \et{0.606}{.006} & \et{0.729}{.005} & \et{0.718}{.038} & \et{3.076}{.028} & \et{1.887}{.050} \\
        & \textbf{+ DisCoRD (Ours)} & \et{0.382}{.007} & \et{0.590}{.007} & \et{0.715}{.004} & \et{0.541}{.038} & \et{3.260}{.028} & \et{1.928}{.059} \\
    \cline{2-8}
        \addlinespace[0.3em]  
        & MoMask~\cite{guo2024momask} & \ets{0.433}{.007} & \ets{0.656}{.005} & \etb{0.781}{.005} & \et{0.204}{.011} & \etb{2.779}{.022} & \et{1.131}{.043} \\
        & \textbf{+ DisCoRD (Ours)} & \etb{0.434}{.007} & \etb{0.657}{.005} & \ets{0.775}{.004} & \ets{0.169}{.010} & \ets{2.792}{.015} & \et{1.266}{.046} \\
    \bottomrule
    \end{tabular}
    }
    \caption{\textbf{Additional quantitative evaluation} on the HumanML3D and KIT-ML test sets. $\pm$ indicates a 95\% confidence interval. +DisCoRD indicates that the baseline model’s decoder is replaced with our DisCoRD decoder.
    \textbf{Bold} indicates the best result, while \underline{underscore} refers the second best.}

    \label{tab:quantitative_supp}
\end{table*}

\subsection{Performance on Various Tasks. } In
Table~\ref{tab:cospeech_supp}, we present additional evaluation results for co-speech gesture generation. Following~\cite{Talkshow}, we additionally report Diversity, which measures the variance among multiple samples generated from the same condition, and Beat Consistency (BC), which evaluates the synchronization between the generated motion and the corresponding audio. In Table~\ref{tab:dance_supp}, we provide additional evaluation results for music-to-dance generation. Following~\cite{siyao2022bailando}, we report $\mathrm{FID_k}$ to measure differences in kinetic motion features and $\mathrm{FID_g}$ for geometric motion features. Additionally, we include the Beat Align Score (BAS) to assess the synchronization between motion and music. While ~\cite{EDGE_fidcannotcatch} has shown that these metrics are not fully reliable and often fail to align with actual output quality, we include them to follow established conventions.

% \vspace{-1.0em}
\begin{table*}[t]
    \centering
    \scalebox{0.95}{
    \begin{tabular}{l c c }
    \toprule
    Methods & Diversity $\uparrow$ & BC $\rightarrow$ (0.868) \\
    \hline   
        \addlinespace[0.2em] 
        TalkSHOW~\cite{Talkshow}& 0.821& \textbf{0.872}\\
        \textbf{+DisCoRD(Ours)} & \textbf{0.919}& 0.876\\
    \midrule
         ProbTalk~\cite{probtalk} & 0.259&0.795\\
         \textbf{+DisCoRD(Ours)}& \textbf{0.331}&\textbf{0.866}\\
    \bottomrule
    \end{tabular}
    }
    \caption{\textbf{Additional quantitative results} on each method's SHOW test set. The results demonstrate that our method performs on par with, or surpasses, the baseline models.}
    % \vspace{-2.0em}
    \label{tab:cospeech_supp}
\end{table*}

\begin{table*}[t]
    \centering
    \scalebox{0.95}{
    \begin{tabular}{l c c c}
    \toprule
    Methods & $\mathrm{FID_k}\downarrow$ &  $\mathrm{FID_g}\downarrow$ & BAS $\uparrow$ \\
    \hline 
        \addlinespace[0.2em] 
        Ground Truth & 17.10 & 10.60 & 0.2374 \\
    \midrule
        TM2D~\cite{gong2023tm2d} & \textbf{19.01} & \textbf{20.09} & 0.2049 \\
        \textbf{+DisCoRD(Ours)} & 23.98 & 88.74 & \textbf{0.2190} \\
    \bottomrule
    \end{tabular}
    }
    \caption{\textbf{Additional quantitative results} on the AIST++ test set. The results demonstrate that, although our method shows performance degradation on $\mathrm{FID_k}$ and $\mathrm{FID_g}$, which are known to be unreliable, it achieves improvement in the Beat Align Score.}
    % \vspace{-2.0em}
    \label{tab:dance_supp}
\end{table*}

\section{Additional Qualitative Results}
\label{sec:additional_visualizations}
\subsection{Motion visualization. }
In Figure~\ref{fig:qualitative} and Figure~\ref{fig:supp_qual}, we present qualitative comparisons between our model and other leading approaches.
In Figure~\ref{fig:supp_qual_2}, we additionally display more qualitative results of our method. We observed that our method effectively follows the text prompts while maintaining naturalness in the generated outputs. Again, we highly recommend viewing the accompanying video, as static images are insufficient to fully convey the intricacies of motion.

\begin{figure*}[t] \centering
    \includegraphics[width=0.48\textwidth]{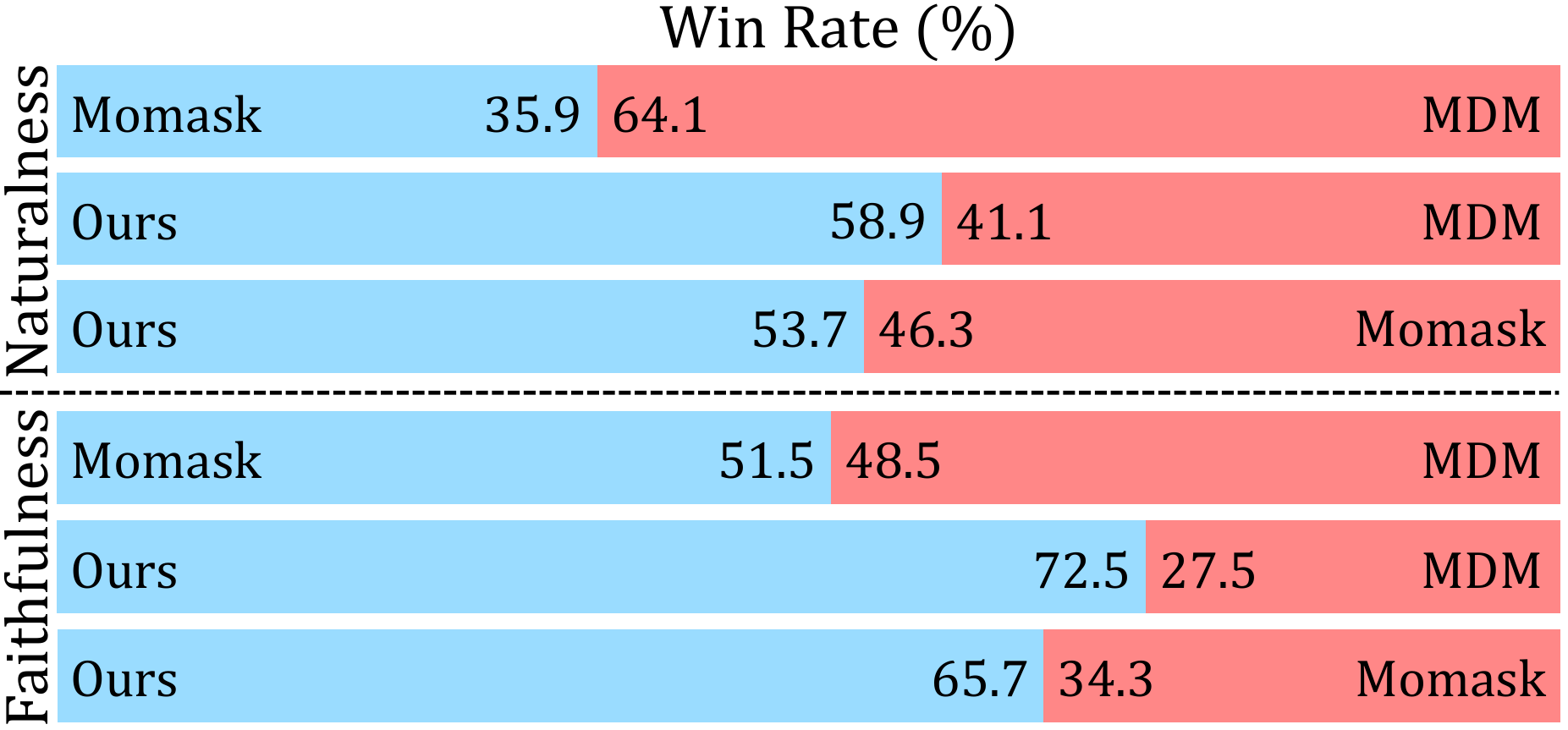}
    \caption{\textbf{User study results on the HumanML3D dataset.} Each bar represents a comparison between two models, with win rates depicted in blue and loss rates in red, evaluated based on naturalness and faithfulness.}
    \vspace{-1.3em}
    \label{fig:user_study}
\end{figure*}
\subsection{User preference study details.} We conduct two user studies to (1) validate our motivation and method effectiveness and (2) evaluate how well sJPE aligns with human perception. The first study, shown in Figure~\ref{fig:user_study}, indicates that the discrete model Momask outperforms the continuous model MDM in faithfulness but lags in naturalness. In contrast, DisCoRD surpasses both, demonstrating its ability to generate motion that is both natural and faithful. In the second study, we find that sJPE exhibits 2.7 times higher correlation with human preference for naturalness compared to MPJPE, highlighting its effectiveness in evaluating sample-wise motion naturalness. Participants were guided to evaluate both faithfulness and naturalness, as shown in  Figure~\ref{fig:survey_1}. Given two motion videos generated by two different models on the same prompt, participants were asked to choose a better one in terms of faithfulness and naturalness, as shown in  Figure~\ref{fig:survey_1_q}. Total 41 participants participated in this user study.  

\begin{figure*}[th] \centering
    \includegraphics[width=0.8\textwidth]{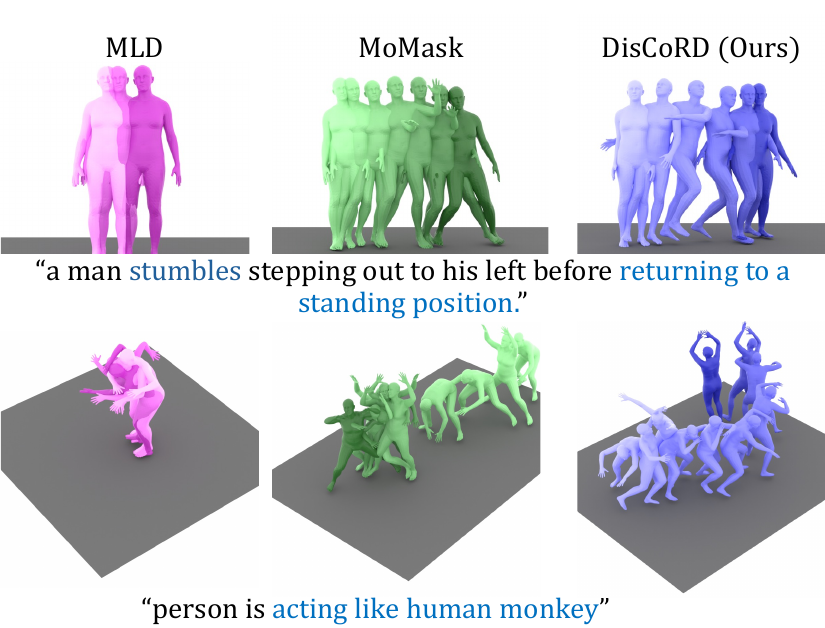}
    \caption{\textbf{Qualitative comparisons} on the test set of HumanML3D.} \label{fig:qualitative}
    % \vspace{-1.3em}
\end{figure*}

\begin{figure*}[th] \centering
    \includegraphics[width=0.8\textwidth]{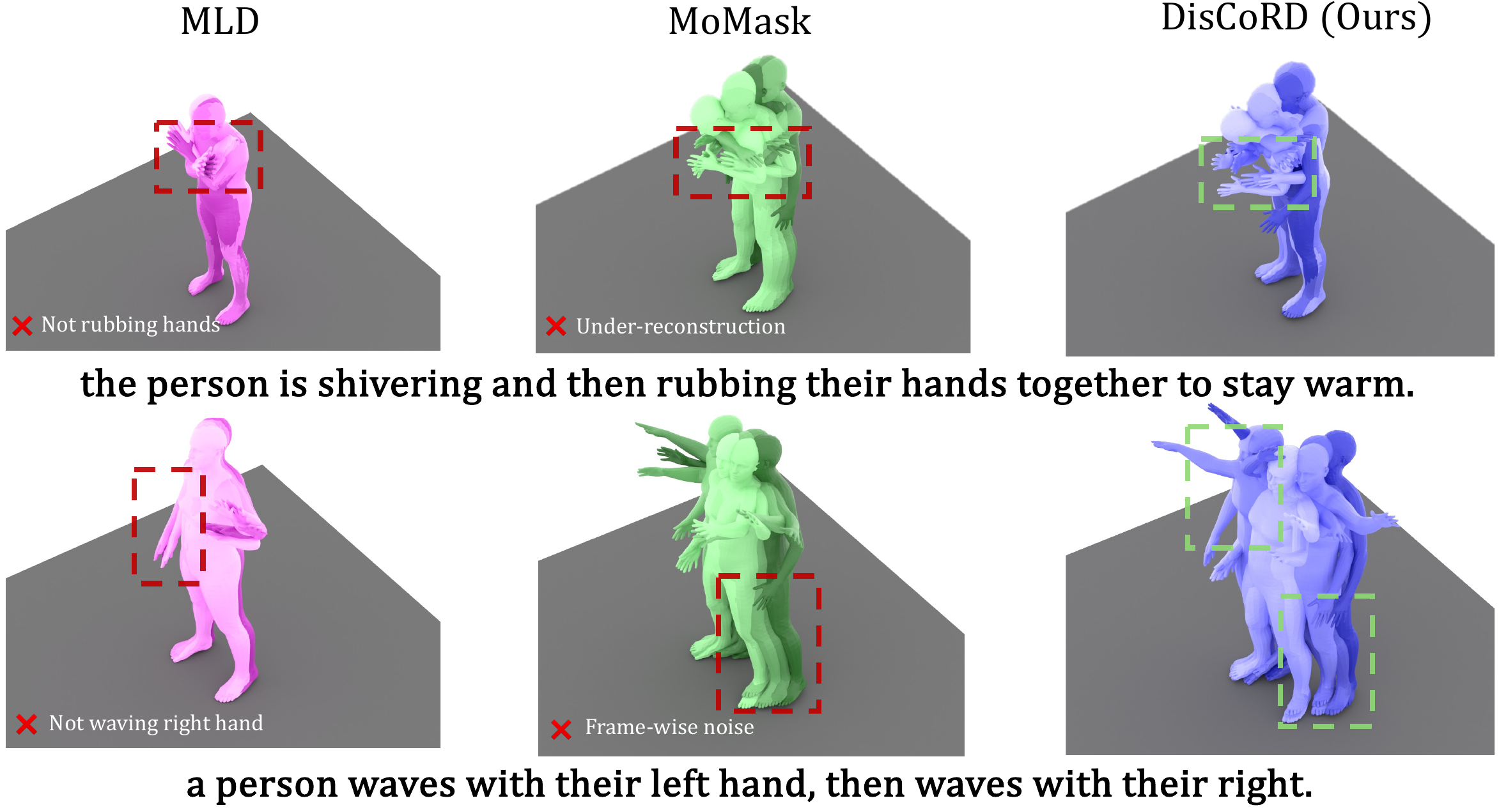}
    \caption{\textbf{Additional qualitative comparisons} on the HumanML3D test set. The continuous method, MLD, often fails to perfectly align with the text consistently, while the discrete method, MoMask, exhibits issues such as under-reconstruction, resulting in minimal hand movement, or unnatural leg jitter caused by frame-wise noise.} \label{fig:supp_qual}
\end{figure*}

\begin{figure*}[h] \centering
\includegraphics[width=0.6\textwidth]{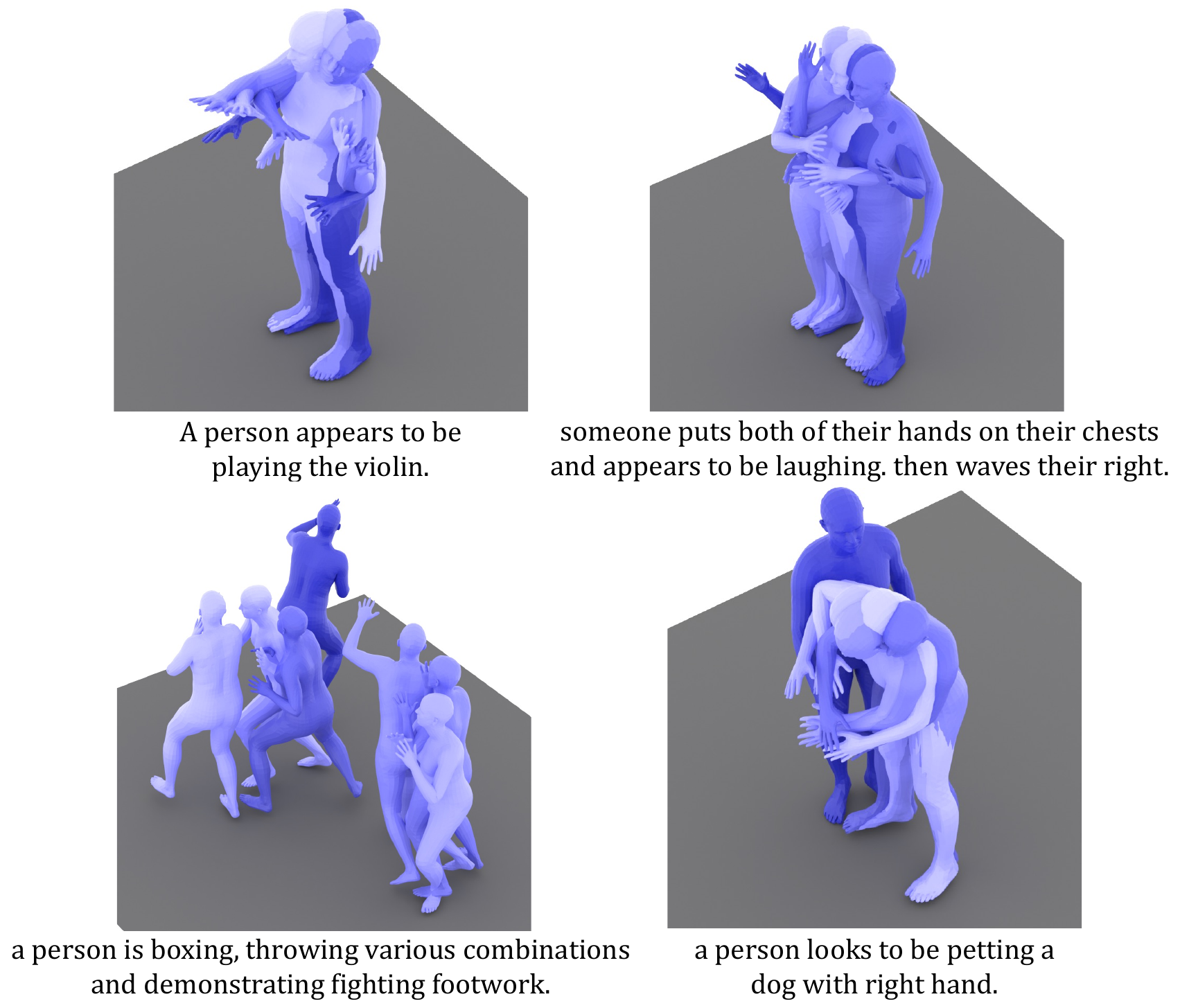}
    \caption{\textbf{Additional qualitative results }of our method on the HumanML3D test set.} \label{fig:supp_qual_2}
\end{figure*}

\begin{table*}[t]
\centering
\small
\setlength{\tabcolsep}{4.5pt}
\vspace{-0.3em}
\resizebox{0.7\linewidth}{!}{
\begin{tabular}{l c c c c c}
\toprule
\addlinespace[-0.01em]
% \multirow{2}{*}{Method} & \multicolumn{4}{c}{Sampling Steps} \\
% \cline{2-5}

\multirow{2}{*}{Method} & \multicolumn{3}{c}{Sampling Steps} \\
\cline{2-5}
\addlinespace[0.2em]
      & 2 & 16 & 50 & 100 \\
\addlinespace[-0.3em]
\midrule
DDPM~(DDIM sample)
 & 0.055 / 0.007s & 0.044 / 0.087s & 0.038 / 0.331s & 0.035 / 0.689s \\
\addlinespace[0.15em]
Linear SDE
      & 8.998 / 0.024s & 0.047 / 0.157s & 0.033 / 0.472s & 0.032 / 0.955s \\
\addlinespace[0.15em]
Ours
      & 0.034 / 0.016s & \textbf{0.032 / 0.221s} & 0.032 / 0.703s & 0.032  / 1.426s \\
\bottomrule
\end{tabular}
}
\caption{FID and decoding time (s) for different sampling steps $(\downarrow)$. The original \textbf{MoMask has a decoding time of 0.244 seconds}. We trained a diffusion decoder (DDPM) using the same architecture as our rectified flow decoder for a fair comparison. For the Linear SDE variant, we replaced only the sampler in our model with the Euler-Maruyama sampler, keeping all other components identical.}
\label{tab:speed_vs_steps}
\end{table*}

\begin{figure*}[h] \centering
    \includegraphics[width=0.8\textwidth]{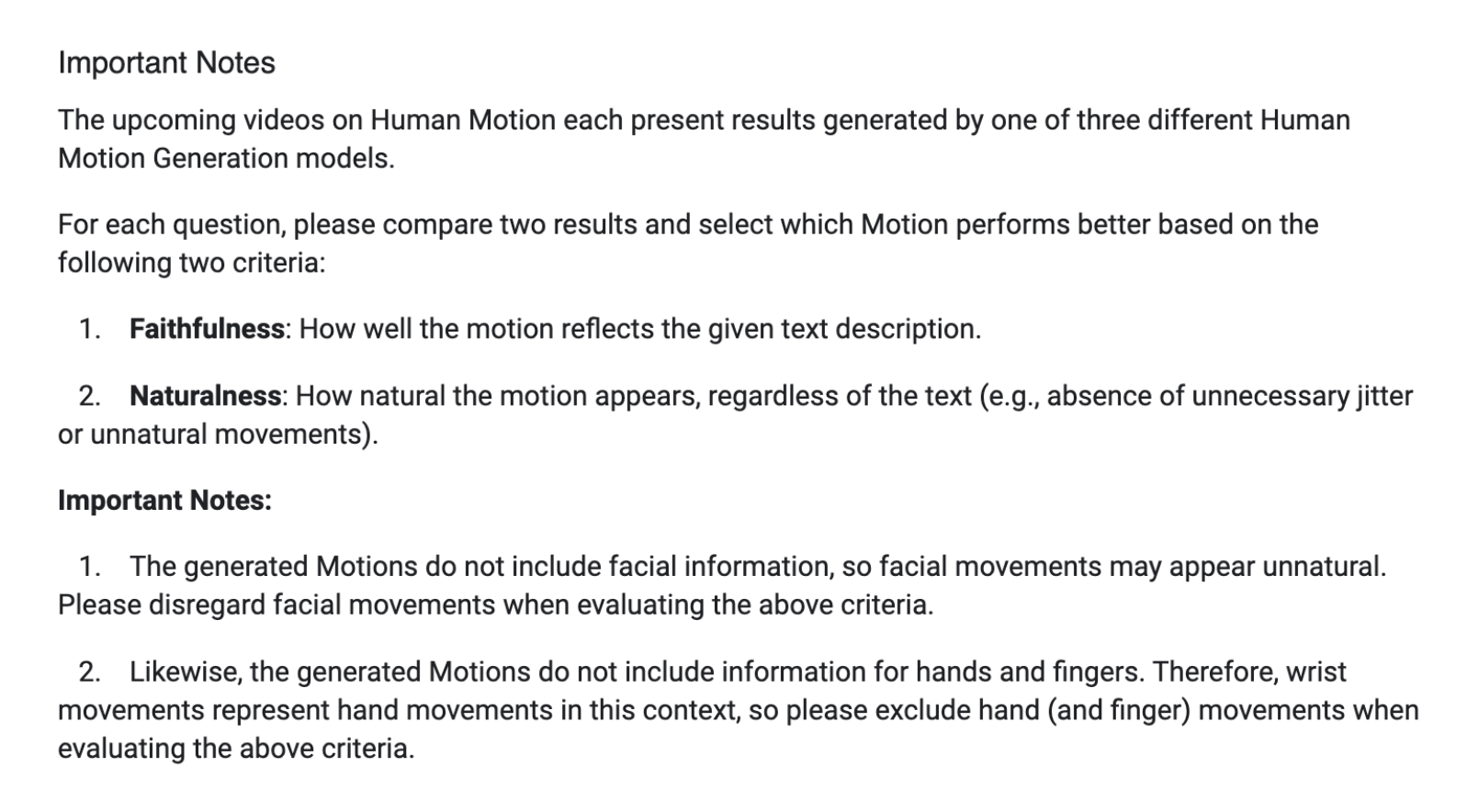}
    \caption{\textbf{Guidelines for user study in the Main paper:} participants were asked to evaluate Faithfulness and Naturalness, excluding hand and facial movements that are not included in HumanML3D.} \label{fig:survey_1}
\end{figure*}

\begin{figure*}[h] \centering
    \includegraphics[width=0.8\textwidth]{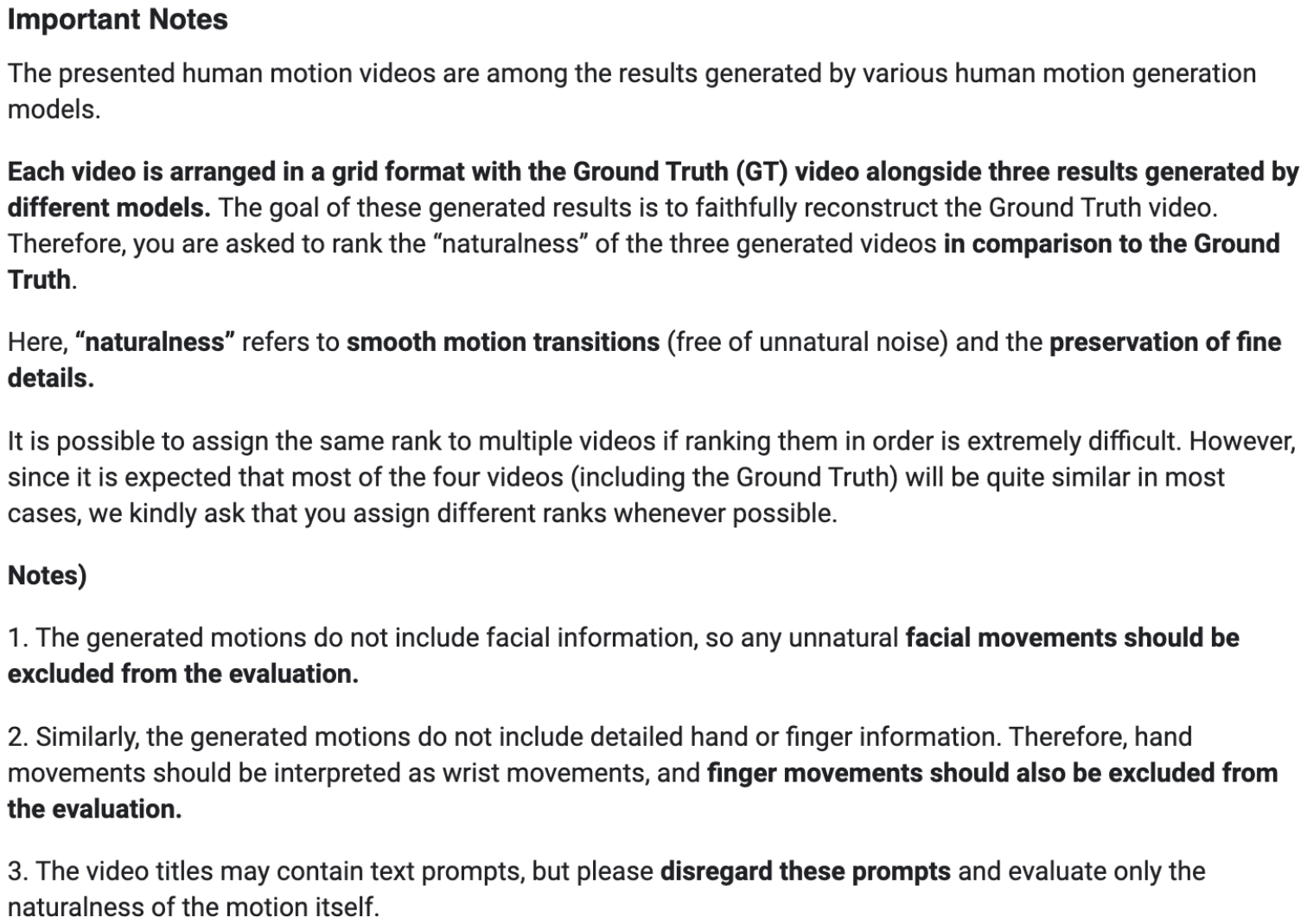}
    \caption{\textbf{Guidelines for User Study in the Supplementary:} Participants were asked to evaluate Naturalness, excluding hand and facial movements that are not included in HumanML3D.} \label{fig:survey_2}
\end{figure*}

\begin{figure*}[h] \centering
    \includegraphics[width=0.7\textwidth]{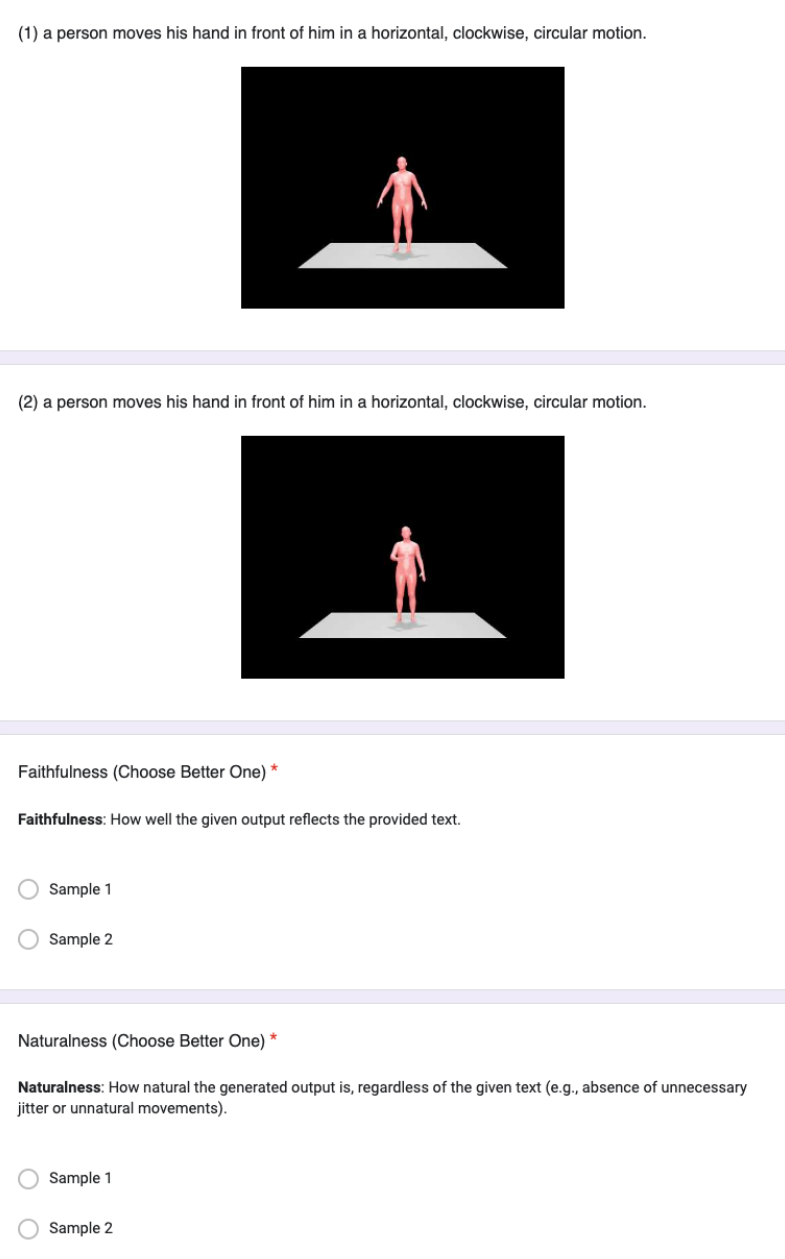}
    \caption{\textbf{User evaluation interface} for the user study in the Main paper: participants were presented with two randomly selected videos and asked to choose the better sample in terms of faithfulness and naturalness.} \label{fig:survey_1_q}
\end{figure*}

\begin{figure*}[h] \centering
    \includegraphics[width=0.7\textwidth]{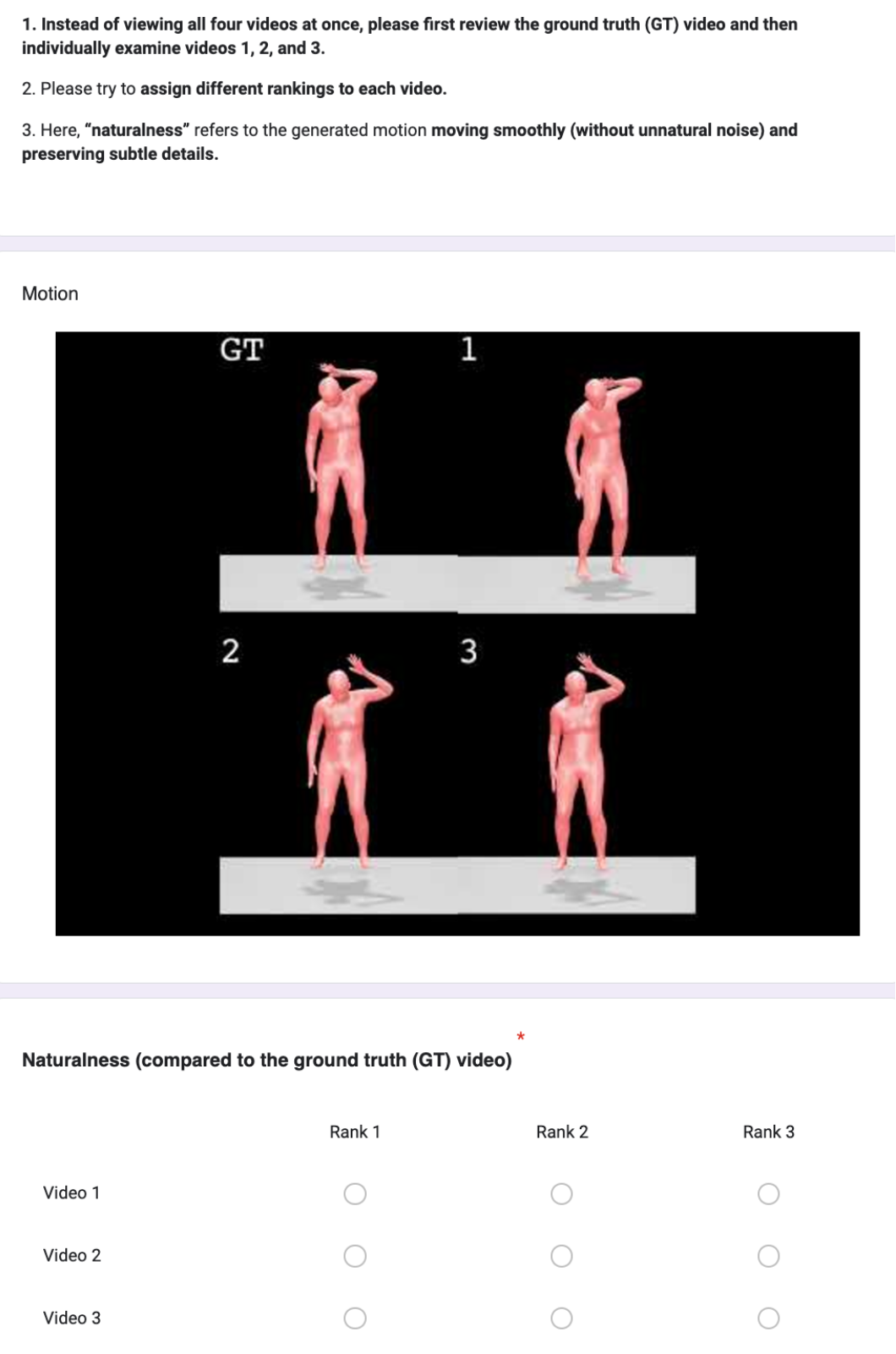}
    \caption{\textbf{User evaluation interface} for the user study in the supplementary: participants were presented with a grid layout containing the GT video and three generated videos. Using the GT video as the upper bound, they were asked to rank the three generated videos in terms of naturalness.} \label{fig:survey_2_q}
    % \vspace{-10em}
\end{figure*}

% \section{Rationale}
% \label{sec:rationale}

% % 
% Having the supplementary compiled together with the main paper means that:
% % 
% \begin{itemize}
% \item The supplementary can back-reference sections of the main paper, for example, we can refer to \cref{sec:intro};
% \item The main paper can forward reference sub-sections within the supplementary explicitly (e.g. referring to a particular experiment); 
% \item When submitted to arXiv, the supplementary will already included at the end of the paper.
% \end{itemize}
% % 
% To split the supplementary pages from the main paper, you can use \href{https://support.apple.com/en-ca/guide/preview/prvw11793/mac#:~:text=Delete%20a%20page%20from%20a,or%20choose%20Edit%20%3E%20Delete).}{Preview (on macOS)}, \href{https://www.adobe.com/acrobat/how-to/delete-pages-from-pdf.html#:~:text=Choose%20%E2%80%9CTools%E2%80%9D%20%3E%20%E2%80%9COrganize,or%20pages%20from%20the%20file.}{Adobe Acrobat} (on all OSs), as well as \href{https://superuser.com/questions/517986/is-it-possible-to-delete-some-pages-of-a-pdf-document}{command line tools}.